\documentclass[runningheads]{llncs}

\usepackage{eccv}

\usepackage{eccvabbrv}

\usepackage{graphicx}
\usepackage{booktabs}

\usepackage[accsupp]{axessibility}  

\usepackage[export]{adjustbox}
\usepackage{overpic}
\usepackage{tikz}
\usepackage{multirow}
\usepackage{stfloats}
\RequirePackage[inline]{enumitem}
\setlist[itemize]{noitemsep,leftmargin=*,topsep=0em}
\setlist[enumerate]{noitemsep,leftmargin=*,topsep=0em}

\DeclareMathOperator*{\argmax}{argmax}

\DeclareMathOperator*{\argtopM}{argtopM}
\newcommand{\shorttextrightarrow}{\clipbox*{{.25\width} 0pt {\width} {\height}} \textrightarrow}

\usepackage{hyperref}

\usepackage{orcidlink}

\begin{document}

\title{SparseRadNet: Sparse Perception Neural Network on Subsampled Radar Data} 

\titlerunning{SparseRadNet: Sparse Perception NN on Subsampled Radar Data}

\author{Jialong Wu\inst{1, 2} \and
Mirko Meuter\inst{2} \and
Markus Schoeler\inst{2} \and 
Matthias Rottmann\inst{1}}

\authorrunning{J.~Wu et al.}

\institute{University of Wuppertal, Wuppertal, Germany\\ \email{jialong.wu@uni-wuppertal.de} \and
Aptiv Services Deutschland GmbH, Wuppertal, Germany}

\maketitle

\begin{abstract}
Radar-based perception has gained increasing attention in autonomous driving, yet the inherent sparsity of radars poses challenges. Radar raw data often contains excessive noise, whereas radar point clouds retain only limited information. In this work, we holistically treat the sparse nature of radar data by introducing an adaptive subsampling method together with a tailored network architecture that exploits the sparsity patterns to discover global and local dependencies in the radar signal. Our subsampling module selects a subset of pixels from range-doppler (RD) spectra that contribute most to the downstream perception tasks. To improve the feature extraction on sparse subsampled data, we propose a new way of applying graph neural networks on radar data and design a novel two-branch backbone to capture both global and local neighbor information. An attentive fusion module is applied to combine features from both branches. Experiments on the RADIal dataset show that our SparseRadNet exceeds state-of-the-art (SOTA) performance in object detection and achieves close to SOTA accuracy in freespace segmentation, meanwhile using sparse subsampled input data.
\end{abstract}

\section{Introduction}
\label{sec:intro}

Environmental perception is vital in autonomous driving. Cameras, LiDARs, and radars capture surrounding data, supporting perception tasks like object detection and segmentation. Unlike vision-based sensors, radars rely on electromagnetic wave reflection, making them more robust to different illumination and weather conditions \cite{RADIATEdataset}. The Doppler effect enables radars to measure object radial velocity, aiding in motion planning \cite{mmwaveReview}. Current industry trends favor multiple-input multiple-output (MIMO) radar systems \cite{mimoRef2}, where multiple transmitting (Tx) and receiving antennas (Rx) form a virtual array to increase spatial resolution and mitigate the sparsity issue of radar data to some extent.

\begin{figure}[htbp] \centering
\includegraphics[width=\textwidth]{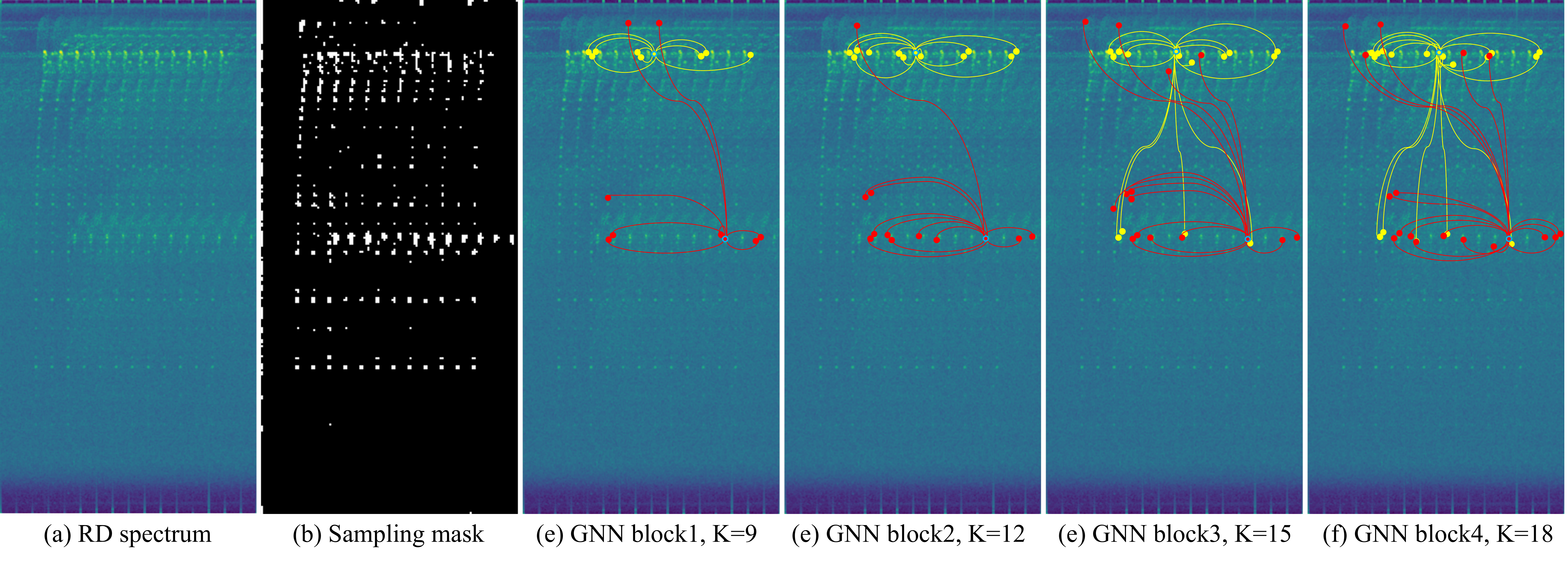}
    \caption{Examples of the (a) RD spectrum, (b) dynamic sampling mask, and (c)-(f) dynamic edges built in our GNN blocks with different number of feature space neighbors.} 
    \label{fig:sec1_example_mask}
\end{figure}

In this work, we study radar-only perception. Radar data can be categorized into radar raw data and point clouds, with the latter obtained after applying the constant false alarm rate (CFAR) algorithm. Radar point clouds are more interpretable by humans and can often be directly input into existing 3D perception models with minor modifications \cite{RadarPillars, Pointillism}. However, this simplicity comes at the cost of losing valuable information present in raw data after the CFAR algorithm. In recent years, there has been a growing preference for perception based on radar raw data, which can have various input forms such as range-angle-Doppler (RAD) tensor \cite{RADDet}, range-Doppler (RD) spectrum \cite{FFT-RadNet}, analog-to-digital converter (ADC) data \cite{ADCNet}, etc.

Existing radar perception models primarily perform convolutional neural networks (CNNs) on radar raw data, while the sparse nature of raw data is rarely explored. Since radar signals are reflection-based, a large portion of raw data pixels contain only noise rather than meaningful reflection from objects, different from the background environment in camera images or the empty spaces in LiDAR point clouds. This also happens to high definition (HD) radars, as shown in Fig. \hyperref[fig:sec1_example_mask]{1a}. Therefore, directly applying dense CNNs to raw data is inefficient.

In this work, we focus on the sparsity issue of radar raw data, aiming to filter out irrelevant noise and process selected crucial information using a suitable network architecture. Huijben \etal~\cite{DPS} introduce deep probabilistic subsampling (DPS) to select a subset of input pixels. DPS is end-to-end trained with the gradient from downstream tasks, \eg image classification or signal reconstruction \cite{ADPS}. We develop a DPS-like method for radar perception scenarios. Instead of using a static sampling mask for any input \cite{DPS}, we dynamically generate a sampling mask for each radar frame. The reason is that for tasks like image classification, important parts are usually located at the center of an image, which is not the case in object detection. Our radar signal subsampling method samples a given number of pixels that contribute most to perception. It masks out noisy pixels without target reflection and allows for data compression. Our subsampling method selects a more representative subset compared to traditional sub-selection algorithms like CA-CFAR \cite{CACFAR} or constant threshold \cite{constThresh}.

On the other hand, applying dense convolution on the subsampled data causes unnecessary computational costs. To concentrate on valid parts of the sparse data, we propose a new approach of applying graph neural networks (GNNs) to radar data. Unlike previous radar GNN models that build static graphs on either radar point clouds \cite{KPConvPillars} or on all raw data pixels \cite{RadGCN}, we only take subsampled pixels as nodes and build dynamic edges according to feature space distances. Our approach does not suffer from the information loss in point cloud-based models, and it also avoids the generation of excessively large graphs. Sparse CNNs (SCNNs) are highly efficient in processing sparse data. We design a novel two-branch backbone that combines GNNs and SCNNs to capture both global and local neighbor information. We also apply an attentive fusion module to fuse global and local features, whose context window is adjusted based on prior knowledge. Our SparseRadNet is evaluated on the RADIal dataset \cite{FFT-RadNet}. Given an RD spectrum, we subsample it to only 3\% of the pixels as model input. Our model exceeds SOTA performance in object detection and achieves close to SOTA accuracy in freespace segmentation, meanwhile using significantly subsampled input data.

Our main contributions can be summarized as follows:
\begin{itemize}
    \item We introduce a deep radar subsampling module for radar perception scenarios, compressing radar raw data and removing noise.
    \item We propose a new architecture that exploits the sparsity of the subsampled signals. Our two-branch backbone combines GNNs and SCNNs to leverage both global and local neighbor information and uses an attentive fusion module to fuse features from different aspects.
    \item We outperform all previous models in object detection on the RADIal dataset. Ablation studies show the synergies of the different modules in our network.

\end{itemize}

\section{Related Work} \label{sec:related_works}

\subsection{Radar Perception Model}

Radar perception models can be categorized based on the data format. Radar signal processing starts with collected ADC data processed by the range and Doppler DFT (Discrete Fourier Transform) resulting in an RD spectrum \cite{review}. Then a third azimuth DFT is applied to generate a RAD tensor. Alternatively, the CFAR algorithm can be executed prior to the azimuth DFT to identify target responses and generate a radar point cloud.

\textbf{Point cloud.} Radar point clouds can be adapted as input to PointNet-like neural networks \cite{PointNet++} with minor adjustments \cite{2DProject, Pointillism, PointNetIDP}. To increase density, methods usually accumulate points \cite{RadarPillars} or stack feature maps from multiple frames \cite{RadarMFNet}. Schumann \etal~\cite{DynStatic} handle moving and static points separately with recurrent layers and stacked histograms. Li \etal~\cite{TRL} introduce temporal relation layers to capture object-level relations. RTCNet \cite{RTCNet} extends point-wise features with corresponding vectors in the RAD tensor. KPConvPillars \cite{KPConvPillars} proposes to use GNN and kernel point convolutions \cite{KPConv} to extract point neighbor features.

\textbf{RAD tensor.} RAD cubes contain rich information but are hard to work with due to their large 3D shapes. RADDet \cite{RADDet} treats the Doppler dimension as the feature channel and performs 2D CNNs. A more popular approach is respectively summing signals power along three dimensions to create RA, RD, and AD maps \cite{3ViewConcat}. RAMP-CNN \cite{RAMP-CNN} extends this approach to multiple frames and its multi-view fusion module facilitates predictions on the RA view. TMVA-Net \cite{MVRSS} applies atrous spatial pyramidal pooling \cite{ASPP} and devises a set of losses to train multi-view heads coherently. Zhang \etal~\cite{PeakConv} propose the PeakConv operation, which imitates the receptive field of CFAR. TransRadar \cite{TransRadar} introduces the adaptive-directional attention block to enhance the feature extraction.

\textbf{RA or RD spectrum.} RA and RD spectra are more compact representations and require only two DFTs in pre-processing. RODNet \cite{RODNet} introduces the M-Net to encode radar signals and the temporal deformable convolution to utilize temporal information. GTR-Net \cite{RadGCN} builds a dense GNN on RA spectra, treating all pixels as nodes, and edges connect direct spatial neighbors. Dong \etal~\cite{RadarUncer} investigate aleatoric uncertainty in radar perception. Radatron \cite{Radatron} designs a two stream model for fusing HD and LD radar data. Zhang \etal~\cite{PhaseNorm} take a U-Net \cite{UNet} architecture on RD spectra and apply phase-normalization to dampen the effect of phase shift. DAROD \cite{DAROD} adopts a lightweight Faster-RCNN \cite{FasterR-CNN} with customized region proposal network. These models mostly apply dense CNN layers to radar spectra, while the sparse and noisy nature of raw data is rarely investigated. We propose a network architecture specifically tailored to discover hidden correlations between sparse radar signals.

The RADIal dataset was collected by a frequency modulated continuous wave (FMCW) radar \cite{FMCW}, where multiple transmitters (Tx) emit the signal with a fixed Doppler shift. As a result, a target reflection appears $N_{\text{Tx}}$ times on the RD spectrum in equidistant Doppler bins. FFT-RadNet \cite{FFT-RadNet} serves as the baseline for our model. It introduces the MIMO pre-encoder to reorganize the repetition of signals and the range-angle decoder to transform RD features to RA space. Jin \etal~\cite{XModSup} apply cross-modal supervision to reduce the burden and errors in labelling. T-FFTRadNet \cite{T-FFTRadNet} directly leverages ADC data, designs the Fourier-Net to mimic DFTs, and uses Swin Transformer \cite{Swin} as the backbone. ADCNet \cite{ADCNet} also operates on ADC data, employing guided pre-training and perturbed initialization to refine trainable DFT layers. These methods aim to obtain a wider range of information from raw ADC. In contrast, we introduce a subsampling method to address the sparsity by condensing radar spectra to considerably less data. This prevents the network from searching for patterns in the noise.

\subsection{Graph Neural Networks}

GNNs have achieved great success in LiDAR \cite{Point-GNN, SVGA-Net} and image \cite{ViG, MobileViG} object detection. However, previous GNN-based radar models have not effectively exploited radar data. They either build graphs on radar point clouds \cite{KPConvPillars, RadarGNN}, which heavily rely on CFAR, or densely take all raw data pixels as nodes \cite{RadGCN}, which is computationally expensive. Moreover, these works build static GNNs, where edges connect nearest spatial neighbors based on point locations or pixel coordinates. In our approach, we select only subsampled RD spectrum pixels as graph nodes, and we build dynamic graphs based on feature distances. Our approach is better suited for FMCW radar spectra, given the aforementioned repetition of object signatures. Our method circumvents the information loss caused by CFAR and requires less computational overhead compared to dense graphs.

\begin{figure*}[tbp] \centering

    \includegraphics[width=\textwidth]{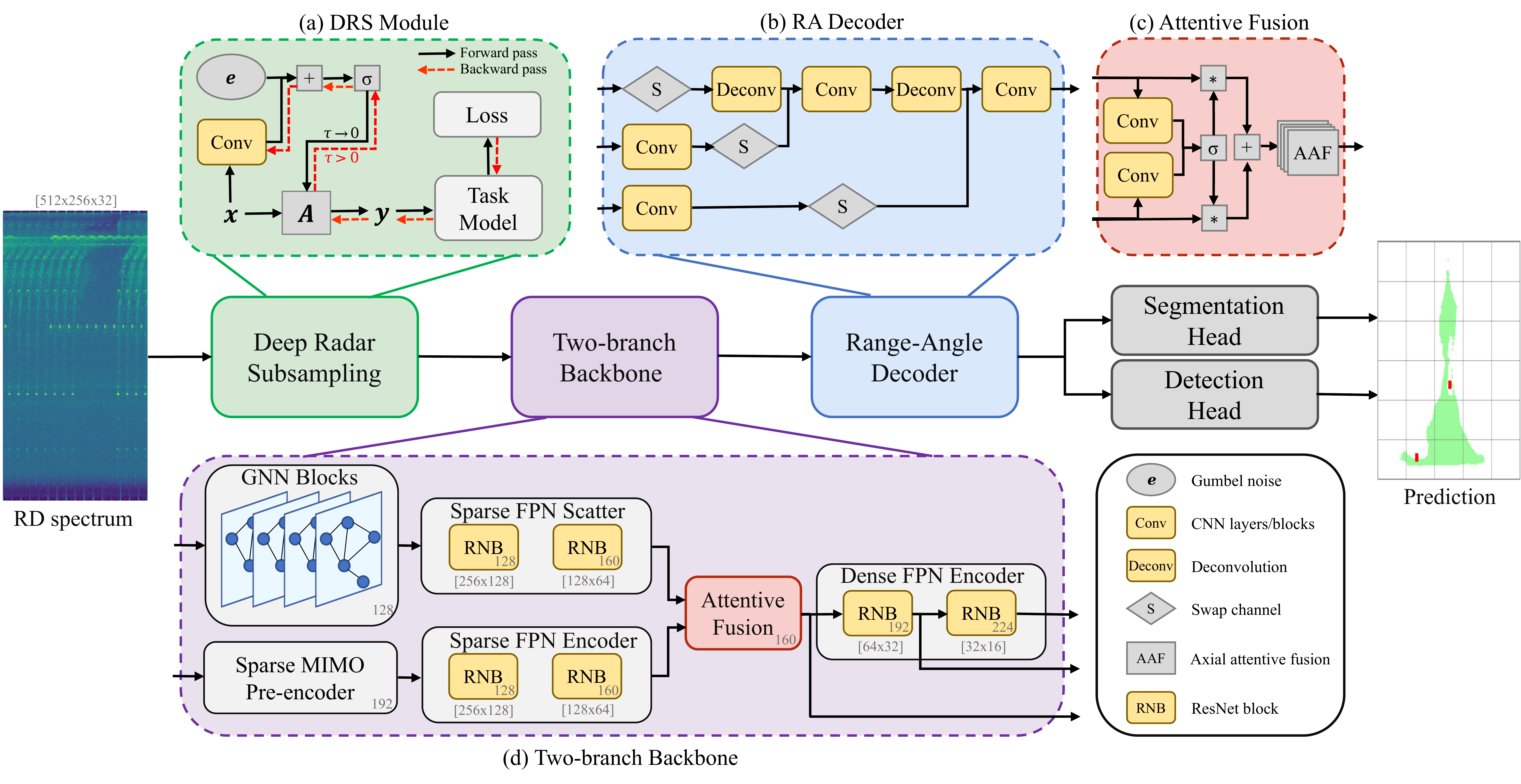}
    \caption{Overview of SparseRadNet. Our model takes RD spectra as input and subsamples them using the deep radar subsampling module. Then the two-branch backbone and attentive fusion enhance the feature extraction from the subsampled data. The range-angle decoder converts the RD view feature map to the RA space. Finally, two perception heads generate object detection and free driving space segmentation results.} \label{fig:sec3_architecture}
\end{figure*}

\section{SparseRadNet architecture} \label{sec:method}

Figure \ref{fig:sec3_architecture} shows the architecture of our SparseRadNet, which has four modules: 
\begin{enumerate*}[label=(\roman*)]
\item the deep radar subsampling module for selecting important parts from the input RD spectrum (\cref{sec:3.1});
\item the two-branch backbone with an attentive fusion module for capturing neighbor information (\cref{sec:3.2});
\item the range-angle decoder for transforming from RD to RA view (\cref{sec:3.3}); and 
\item two output heads for object detection and freespace segmentation (\cref{sec:3.4}).
\end{enumerate*}

\subsection{Deep Radar Subsampling Module} \label{sec:3.1}

Radar raw data contains rich information, yet its sparsity poses challenges to radar perception. To address this, we introduce a DPS-like method \cite{DPS} for radar perception, which we refer to as the deep radar subsampling (DRS) module, a sketch shown in Fig. \hyperref[fig:sec3_architecture]{2a}. 
This module selects an important subset of pixels that contribute most to perception, while masking out irrelevant noise. Identifying important parts from sparse radar data enables the task model to concentrate on valid target reflections and allows for data compression.

Initially, RD spectra are complex tensors with shape $\mathbb{C}^{H \times W \times N_{\text{Rx}}}$, where $N_{\text{Rx}}$ is the number of receivers. Complex numbers are decomposed into real and imaginary parts and taken as our initial input $\bold{x} \in \mathbb{R}^{H \times W \times 2N_{\text{Rx}}}$. Let $\bold{A} \in \left \{ 0, 1 \right \}^{H\times W}$ be a binary sampling mask (see Fig. \hyperref[fig:sec1_example_mask]{1b} for an example), performing an element-wise multiplication with the input in terms of $\bold{y}_{h, w, c} = \bold{A}_{h, w} \cdot \bold{x}_{h, w, c}$, we obtain a sparse representation $\bold{y} \in \mathbb{R}^{H \times W \times 2N_{\text{Rx}}}$. The number of selected elements $M$ in mask $\bold{A}$ is predefined and significantly fewer than the total spatial cells $N = H \times W$ ($M \ll N$). In the original DPS method, the binary mask $\bold{A}$ is obtained from a subsampling distribution $\bold{A}_\Phi$, which is parameterized by a set of learnable parameters $\Phi \in \mathbb{R}^{H \times W}$. Parameters are trained by the downstream task's loss, and once trained, all input frames share the same $\Phi$.  In contrast, we generate the sampling mask based on the input to deal with dynamic scenarios in radar perception. We utilize neural networks to operate on the initial input $\bold{x}$ and consider the output as unnormalized logits $\bold{Z}_\theta(\bold{x}) \in \mathbb{R}^{H \times W}$, with $\theta$ representing the parameters of CNN layers.

We frame the problem as sampling from a categorical distribution. Instead of applying a softmax function on $\bold{Z}_\theta(\bold{x})$ to get a probability distribution $\bold{A}_\theta(\bold{x})$, we use the Gumbel-Softmax method \cite{GumbelSoftmax} to sample $M$ pixels directly from the unnormalized logits. The Gumbel-Max trick \cite{GumbelTrick} offers an efficient way of categorical sampling, and Gumbel-Softmax extends this method as a reparameterization trick to make the sampling process differentiable. Each element $z_{h, w}$ in $\bold{Z}_\theta(\bold{x})$ is firstly perturbed with Gumbel noise $e_{h, w} \sim \mathrm{Gumbel}(0, 1)$. The perturbation moves the non-differentiable stochastic node to the edge of the computational graph, which enables gradient backpropagation to our CNN layers. Then applying the $\argmax$ operation to the perturbed logits could generate one sample. This process equals drawing a sample (category) from the categorical distribution $\bold{A}_\theta(\bold{x})$. We define $\bold{r} = \left \{ \bold{r}_1 , ..., \bold{r}_m \right \} \subset \mathbb{N}^{2} $ as $M$ sampled pixel indices. Following the DPS-topK pattern \cite{DPS}, we take an $\argtopM$ operation to simultaneously sample $M$ pixels without replacement:

\begin{equation}
    \bold{r} = \argtopM_{1 \leq h \leq H \atop 1 \leq w \leq W } \left \{ z_{h, w} + e_{h, w} \right \}.
\end{equation}
The $\argtopM$ operation is again non-differentiable, just like $\argmax$. We could build a binary hard sampling mask $\bold{A}^{hard}$ from $\bold{r}$, but it does not allow parameter training. To solve this, Gumbel-Softmax uses a softmax function as a continuous differentiable approximation. A differentiable soft mask $\bold{A}^{soft}$ is built as

\begin{equation} \label{eq:sec3_Asoft}
    \bold{A}^{soft} =  \sum_{m=1}^{M}  \text{softmax}_{\tau} \left \{ \bold{w}^{m} +  \bold{Z}_\theta(\bold{x}) + \bold{e} \right \},
\end{equation}
where $ w^{m}_{h, w} \in \{-\infty, 0\}$ suppresses those $m-1$ previously drawn samples by setting them to $-\infty$, and $\tau$ denotes the temperature parameter of the softmax. The parameter $\tau$ controls the smoothness of Gumbel-Softmax. When $\tau \to 0$ the soft sampling mask approaches the hard mask. During training we define the sampling mask as

\begin{equation} \label{eq:sec3_overall_mask}
    \bold{A} = \bold{A}^{hard} + \bold{A}^{soft} - \text{detach}(\bold{A}^{soft}),
\end{equation}
where $\text{detach}(\cdot)$ removes a tensor from the computational graph. In the forward pass, mask $\bold{A}$ serves as a binary sampling mask and produces $\bold{y}$ as the sparse input to downstream perception modules (black solid arrow in Fig. \hyperref[fig:sec3_architecture]{2a}). While in the backward pass, gradients propagates to \cref{eq:sec3_overall_mask} where only the second term $\bold{A}^{soft}$ requires gradient. In \cref{eq:sec3_Asoft}, gradients are distributed by the differentiable proxy $\text{softmax}_{\tau}$ to all logits $\bold{Z}_\theta(\bold{x})$, then to our sampling layers' parameters $\theta$ (red dashed arrow in Fig. \hyperref[fig:sec3_architecture]{2a}).

The DRS module is end-to-end trained by gradients from perception losses. With a limited number of samples, it tries to select pixels that have the most significant impact on the perception tasks. After sampling, the sparse input $\bold{y}$ carries sufficient information for perception while reducing the presence of noise.

\subsection{Two-branch Backbone}  \label{sec:3.2}

To fully exploit the sparse subsampled input from the DRS module and avoid unnecessary computations, we introduce the two-branch backbone and the attentive fusion module to enhance the feature extraction. As depicted in Fig. \hyperref[fig:sec3_architecture]{2d} and \hyperref[fig:sec3_architecture]{2c}, the GNN branch captures information of global embedding space neighbors, while the second SCNN branch aggregates features from local neighbors. Then, the attentive fusion module combines global and local features adaptively, with prior radar knowledge imposed on its attentive context window. Finally, The fused feature map is further processed by a dense but lower-resolution feature pyramid network (FPN).

\subsubsection{GNN Branch.}

The GNN branch has multiple GNN blocks and the sparse feature scatter module. We build the GNNs based on the isotropic pattern of ViG \cite{ViG}, where features in between GNN layers keep the same shape and size.

The GNN branch fetches the $M$ subsampled pixels as input, according to indices obtained from the DRS module. Those zeros in $\bold{y}$ resulting from masking are simply ignored. Selected RD pixels carrying $2 \times N_{\text{Rx}}$ channels are firstly expanded by linear layers to $C$ dimensions as $\overline{\bold{y}} \in  \mathbb{R}^{M\times C}$. They are treated as initial node features, and $\mathcal{V}$ denotes graph nodes by convention. To create the graph, for each node $v_i$, we find its $K$ nearest neighbors in the embedding space and connect them, resulting in graph edges $\mathcal{E}$. The number of neighbors $K$ increases as we progress through more GNN blocks. Finally, we construct a graph $\mathcal{G}=(\mathcal{V}, \mathcal{E})$ on subsampled radar pixels. Since node features are updated after each GNN block, the neighbors of a node in the embedding space may change greatly between blocks, therefore called dynamic graphs. Dynamic graphs are more suitable for exploiting the phenomenon of recurring object signatures in FMCW radars (an example of connected dynamic neighbors see Fig. \hyperref[fig:sec1_example_mask]{1c-f}), as opposed to static graphs which primarily focus on local structures.

Each GNN block consists of a Grapher module and a feed-forward network (FFN) \cite{ViG}. The Grapher module has one graph convolutional network (GCN) layer with two linear layers before and after respectively. The inclusion of linear layers and the FFN increases feature diversity. The GCN layer has two steps: aggregation and update. Prior to each aggregation step, a dynamic graph $\mathcal{G}$ is constructed based on the current feature state of nodes. In the aggregation step, the features of node neighbors are collected and aggregated by the operation $g (\cdot)$, and the update operation $h (\cdot)$ updates the features of each node through linear layers. The GCN operation $\text{GraphConv}(\cdot)$ on node embedding $\overline{\bold{y}}_{i}$ is defined as:
\begin{equation}
    \overline{\bold{y}}_{i}^\prime = \text{GraphConv}(\overline{\bold{y}}_{i})= h \left ( \overline{\bold{y}}_{i}, g \left ( \overline{\bold{y}}_{i}, \mathcal{E}(v_{i}) \right ), W_{update} \right ),
\end{equation}
where $i = {1, ..., M}$, $ W_{update}$ is the parameters of the update operation, and two operations \cite{groupedConv} are defined as:
\begin{equation}
\begin{aligned}
    & g(\overline{\bold{y}}_{i}) = \text{concat} \left ( \overline{\bold{y}}_{i}, \max \left ( \left \{ \overline{\bold{y}}_{j} - \overline{\bold{y}}_{i} \mid j \in \mathcal{E}(v_{i}) \right \} \right) \right), \\
    & h(\overline{\bold{y}}_{i})= g(\overline{\bold{y}}_{i}) W_{update},
\end{aligned}
\end{equation}
where $h(\cdot)$ is the grouped convolution. Features in the RD view encompass angle information, and grouping into subspaces can be interpreted as dividing quadrants. Overall, the Grapher module extracts node features by:
\begin{equation}
    \text{Grapher}(\overline{\bold{y}}) = \sigma \left ( \text{GraphConv}\left ( \overline{\bold{y}} W_{in} \right ) \right) W_{out} + \overline{\bold{y}},
\end{equation}
where $\sigma$ is the GeLU activation function \cite{GeLU}, $W_{in}$ and $W_{out}$ are parameters of two linear layers, bias terms are omitted for simplicity.

The sparse FPN scatter includes two sparse ResNet blocks (RNB) \cite{ResNet}, mainly aiming to scatter the extracted node features onto the RD map. Each RNB contains multiple residual blocks, where regular SCNN layers handle feature scattering and spatial size reduction, and submanifold SCNNs \cite{submanifold} are applied for efficient feature refinement.

\subsubsection{SCNN Branch.}

Local structures also contain valuable contextual information. We use the SCNN branch to efficiently extract features from local neighbors, as a complement to the GNN branch. It comprises the sparse MIMO pre-encoder and sparse FPN encoder, using the sparse representation $\bold{y}$ as input.

The MIMO pre-encoder is introduced in FFT-RadNet \cite{FFT-RadNet} for reorganizing the repeated object signatures in RD spectra. It utilizes an atrous convolution layer that is customized to the radar sensor. The kernel size is set to $1 \times N_{\text{Tx}}$, and its dilation $\delta = \frac{\Delta \cdot W}{D_{\text{max}}}$ depends on the transmitters' Doppler shift $\Delta$, number of Doppler bins $W$, and maximum Doppler value $D_{\text{max}}$. By this design, the MIMO pre-encoder can de-interleaved the repetition of target responses. Given that our input $\bold{y}$ has only $M$ valid pixels, we employ a sparse atrous convolution to eliminate redundant computation. The sparse FPN encoder shares the same structure as the scatter in the GNN branch. The only difference is that the sparse FPN encoder has more residual blocks to strengthen the local feature extraction.

\subsubsection{Attentive Fusion.}

The GNN and SCNN branches extract features in different aspects. To take full advantage of the two-branch backbone, we employ the attentive fusion module to fuse two feature maps. This module has two parts: spatial attentive addition and axial attentive fusion, as illustrated in  Fig. \hyperref[fig:sec3_architecture]{2c}.

Firstly, we combine feature maps in a fashion of spatial attention \cite{SpatialAtt}. Two convolutional layers operate on feature maps with shape $\mathbb{R}^{H \times W \times C}$ respectively, generating two attention score maps with shape $\mathbb{R}^{H \times W \times 1}$. After concatenating two score maps, the softmax function is applied pixel-wise to produce two weighting factors, indicating the relative importance of each branch's features at that pixel. The combined feature map is a weighted sum of two branches.

Secondly, we align global and local features using axial attention blocks \cite{AxAtt}. The GNN branch concentrates on the affinity of subsampled pixels and basically maintains their initial position in the RD space. As shown in Fig. \hyperref[fig:sec1_example_mask]{1c-f}, this affinity mainly comes from the repeated object signatures. In the SCNN branch, the repeated object signatures are reorganized via the MIMO pre-encoder, where the $1 \times N_{\text{Tx}}$ dilated kernel permutes object signatures within each row. As a result, the global features and local features are not spatially aligned with each other. On the other hand, we also expect our fusion module to consider column-wise dependencies arising from the shape of objects along the range axis. To address both row-wise alignment and column-wise dependencies, our fusion module utilizes axial attention \cite{AxAtt}, which decomposes one attention block into column-wise and row-wise blocks. Each block restricts the context window inside one row or column. It sets the focus for the attention mechanism and fulfills our prior knowledge about radar signals and network structure.

Since the spatial resolution has been reduced to an affordable level and features have been scattered across the map, the dense FPN encoder uses dense RNBs for further feature extraction. Additionally, skip connections provide features with different resolutions to the range-angle decoder, as shown in Fig. \hyperref[fig:sec3_architecture]{2d}.

\subsection{Range-angle Decoder}  \label{sec:3.3}

The range-angle decoder \cite{FFT-RadNet} is designed to transform the RD view feature map to the RA space for final predictions, as illustrated in Fig. \hyperref[fig:sec3_architecture]{2b}. Feature maps from deeper RNBs have lower resolution but contain high-level semantic features, while shallow feature maps contain low-level spatial information. Combining these feature maps enriches the information for the transformation. CNN layers are used to match the channel dimension with the number of angle bins. Then the swap operation switches the feature channel with the Doppler spatial dimension, treated as an RA view. For lower-resolution feature maps, the deconvolution operation is applied to recover the spatial resolution.

\subsection{Multi-task Head}  \label{sec:3.4}

We use the same multi-task head as FFT-RadNet \cite{FFT-RadNet}. The free space segmentation head makes binary predictions about whether a pixel is free or occupied. The detection head predicts the presence of vehicles and range and angle offsets.

\section{Experiments}

We conducted experiments on the RADIal dataset \cite{FFT-RadNet}. The dataset was recorded by an HD FMCW radar, which has $N_{\text{Tx}}=12$ transmitters and $N_{\text{Rx}}=16$ receivers. It provides raw ADC data, enabling the subsequent generation of all other data representations, \eg RD spectra, as outlined in \cref{sec:related_works}. Following the same data split, the dataset is divided into 6231 training samples, 986 validation samples, and 1035 test samples. Labels are generated with the assistance of the camera and LiDAR, followed by manual verification and categorized as ``easy'' or ``hard'' cases. We follow the same criteria to evaluate our model's performance. For object detection, the RADIal dataset labels all types of vehicles as the "Vehicle" class. We compute the average precision (AP), average recall (AR), average F1 score, as well as mean range and angle error, all within a 100 meters range. A prediction is considered a true positive if its Intersection over Union (IoU) exceeds 0.5 with any ground truth bounding box. Since the dataset has no object size labels, a standard vehicle size is adopted with 4 meters in length and 1.8 meters in width. For the free driving space segmentation task, labels are binary maps indicating free or occupied space. We measure the mean IoU (mIoU) over all test frames within the range $[0, 50]$ meters. In \cref{fig:sec4_qualitative}, we present some qualitative results on the RADIal test set, where the predictions are filtered using the same thresholds as in the evaluation. Note that we only evaluate our model on the RADIal dataset, as it is the only dataset providing both HD radar raw data and object detection annotations.

\subsection{Implementation details} \label{sec:4.2}

In the DRS module, we configure the number of sampled pixels as $M=4000$, which constitutes only 3\% of the input. To reduce the memory required during training and lighten the burden on the softmax, we perform average pooling and sample 1000 $2 \times 2$ patches. The temperature parameter $\tau$ of the softmax is set to 4. To speed up the inference, we do not add Gumbel noise in the evaluation mode and only compute the hard mask $\bold{A}^{hard}$ for a forward pass. A sparse backbone cannot provide complete gradient information for the $\text{softmax}_{\tau}$ over all pixels when gradients backpropagate to $\bold{A}^{soft}$, while a dense backbone enables the training of the DRS module. Meanwhile, the MIMO pre-encoder could incorporate prior knowledge about the repeated object signatures into the gradients. Therefore, we pre-train our DRS module by plugging it into the dense baseline model FFT-RadNet \cite{FFT-RadNet}. Then we load and freeze the weights of this module when training our SparseRadNet.

In the GNN branch, we use four GNN blocks. For each node, we find $K=$ 9, 12, 15, and 18 neighbors respectively. The isotropic feature size between these blocks is 128 channels. Our GNN branch is designed to focus on the global embedding space affinity. Therefore, we opt not to add positional encoding to node features, granting the model full freedom in global exploration. The dilation $\delta$ of the sparse MIMO pre-encoder is 16. The first residual block of each sparse RNB uses a regular SCNN layer with kernel size $3\times3$ and stride 2 to reduce the spatial resolution. In the sparse FPN scatter, one submanifold residual block follows it, and in the sparse FPN encoder there are two. In the attentive fusion module, we apply 8 axial attention blocks. In the dense FPN encoder, two RNBs have 4 and 3 dense residual blocks separately.

We use the Adam optimizer with a learning rate of $10^{-4}$ and a decay weight of 0.9 for every 10 epochs. We train for 100 epochs and pick the model that achieves the highest average F1 score on the validation set. More details about the baseline model, loss functions, evaluation metrics and detailed ablation study set-up can be found in the supplement.

\begin{table}[tbp]\centering
    \caption{Object detection performances on the RADIal test set. Methods are categorized by their input type. Evaluation metrics are average precision (AP), average recall (AR), average F1 score (F1), range error in meter (RE), and angle error in degree (AE). ``Overall'' means the whole test set, while ``Easy'' and ``Hard'' are two difficulty subsets. ``*'': image-based evaluation. ``$\dag$'': result reproduced by us. }
    \label{sec4:table_SparseRadNet}
    \resizebox{\textwidth}{!}{
\begin{tabular}{c|c|ccccc|ccccc|ccccc}
\toprule
\hline
\multirow{2}{*}{\begin{tabular}[c]{@{}c@{}}Input\\ Type\end{tabular}} & \multirow{2}{*}{Model} & \multicolumn{5}{c|}{Overall}                          & \multicolumn{5}{c|}{Easy}           & \multicolumn{5}{c}{Hard}            \\
                                                                      &                        & F1$\uparrow$    & AP$\uparrow$    & AR$\uparrow$    & RE$\downarrow$ & AE$\downarrow$ & F1$\uparrow$    & AP$\uparrow$    & AR$\uparrow$    & RE$\downarrow$ & AE$\downarrow$ & F1$\uparrow$    & AP$\uparrow$    & AR$\uparrow$    & RE$\downarrow$ & AE$\downarrow$   \\ 
\hline
\multirow{2}{*}{ADC}                                                  & T-FFTRadNet \cite{T-FFTRadNet} & 87.44          & 88.20 & 86.70 & 0.16 & 0.13          & \multicolumn{5}{c|}{N/A}                         & \multicolumn{5}{c}{N/A}                         \\
                                                                      & ADCNet  \cite{ADCNet} & 91.90          & 95.00 & 89.00 & 0.13 & 0.11          & 96.99 & 96.00 & 98.00 & 0.12        & 0.11       & 81.01 & 91.00 & 73.00 & 0.16       & 0.12       \\ \hline

\midrule
\multirow{6}{*}{RD}                                                   & FFTRadNet \cite{FFT-RadNet} & 88.91          & 96.84 & 82.18 & 0.11 & 0.17          & 94.97 & 98.49 & 91.69 & 0.10        & 0.13       & 76.37 & 92.93 & 64.82 & 0.13       & 0.26       \\
                                                                      & T-FFTRadNet \cite{T-FFTRadNet} & 89.50          & 89.60 & 89.50 & 0.15 & 0.12          & \multicolumn{5}{c|}{N/A}                         & \multicolumn{5}{c}{N/A} \\
                                                                      & Cross Modal DNN* \cite{XModSup}& 89.70          & 96.90 & 83.49 & N/A  & N/A           & 95.18 & 98.61 & 91.98 & N/A  & N/A  & 77.41 & 92.79 & 66.41 & N/A  & N/A  \\
                                                                      & FFT-RadNet\textsuperscript{\textdagger} \cite{FFT-RadNet}& 90.75          & 95.03 & 86.83 & 0.12 & 0.11 & \underline{96.82}    & 97.19 & 96.46 & 0.11 & 0.10 & 78.29          & 90.00 & 69.26 & 0.15 & 0.13 \\
                                                                      & TransRadar\textsuperscript{\textdagger} \cite{TransRadar} & \underline{92.87}    & 94.91 & 90.90 & 0.15 & 0.10 & 96.57          & 96.52 & 96.61 & 0.14 & 0.10 & \textbf{85.68} & 91.59 & 80.49 & 0.17 & 0.11 \\
                                                                      & SparseRadNet (ours)    & \textbf{93.84}          & 96.00 & 91.78 & 0.13 & 0.10 & \textbf{98.29}          & 98.02 & 98.57 & 0.12 & 0.09 & \underline{85.45}          & 91.90 & 79.85 & 0.16 & 0.12 \\ \hline
\bottomrule
\end{tabular}%
    }
\end{table}

\subsection{Comparison with State of the Art}  \label{sec:4.3} 

We compare our SparseRadNet with the state-of-the-art methods on the RADIal test set. Table \ref{sec4:table_SparseRadNet} lists the model performances on the object detection task. The average F1 score is considered the main metric as it provides a trade-off between AP and AR. When reproducing the result for the baseline FFT-RadNet \cite{FFT-RadNet}, we achieved higher scores compared to what was reported in \cite{FFT-RadNet}. Our model ranks 1st place on the object detection task, achieving higher accuracy than the previous best model TransRadar \cite{TransRadar} by 1 point in the overall test set. When examining the two difficulty levels separately, our model outperforms TransRadar by 1.7 points in the easy mode but falls short by around 0.2 points in the hard mode. This difference can be attributed to our DRS module, which selects only 4000 pixels as the sparse input, accounting for only 3\% of the RD spectrum. Some hard cases may not have their object signatures fully selected in sampling, resulting in difficulties in detection. As a result, our model achieves a higher average precision but relatively lower average recall. As mentioned in \cref{sec:4.2}, our DRS module requires full gradients and relies entirely on the pre-training with a dense backbone, which hinders fine-tuning the sampling on sparse models. This constraint represents the main limitation of our DRS module.

\begin{table}[htbp]\centering
    \caption{Freespace segmentation performances on the RADIal test set. The evaluation metric is the mean IoU (mIoU). ``$\dag$'': result reproduced by us. }
    \label{sec4:table_semseg}
    \resizebox{0.4\textwidth}{!}{
\begin{tabular}{c|c|ccc}
\toprule
\hline
\multirow{2}{*}{\begin{tabular}[c]{@{}c@{}}Input\\ Type\end{tabular}} & \multirow{2}{*}{Model} & \multicolumn{3}{c}{mIoU$\uparrow$}                         \\
                                                                      &                        & Overall        & Easy           & Hard           \\
\hline
\multirow{2}{*}{ADC}                                                  & T-FFTRadNet \cite{T-FFTRadNet} & 79.60    & N/A   & N/A   \\
                                                                      & ADCNet  \cite{ADCNet} & 78.59    & 79.63 & 75.90 \\ \hline

\midrule
\multirow{6}{*}{RD}                                                   & FFTRadNet \cite{FFT-RadNet} & 74.00    & 74.60 & 72.30 \\
                                                                      & T-FFTRadNet \cite{T-FFTRadNet} & 80.20    & N/A   & N/A   \\
                                                                      & Cross Modal DNN \cite{XModSup} & \underline{80.40}    & \underline{81.60} & \underline{76.70} \\
                                                                      & FFT-RadNet\textsuperscript{\textdagger} \cite{FFT-RadNet} & 77.37          & 78.38          & 74.78          \\
                                                                      & TransRadar\textsuperscript{\textdagger} \cite{TransRadar} & \textbf{82.27} & \textbf{83.16} & \textbf{80.00} \\
                                                                      & SparseRadNet (ours)    & 78.48    & 79.38          & 76.29          \\ \hline
\bottomrule
\end{tabular}%
    }
\end{table}

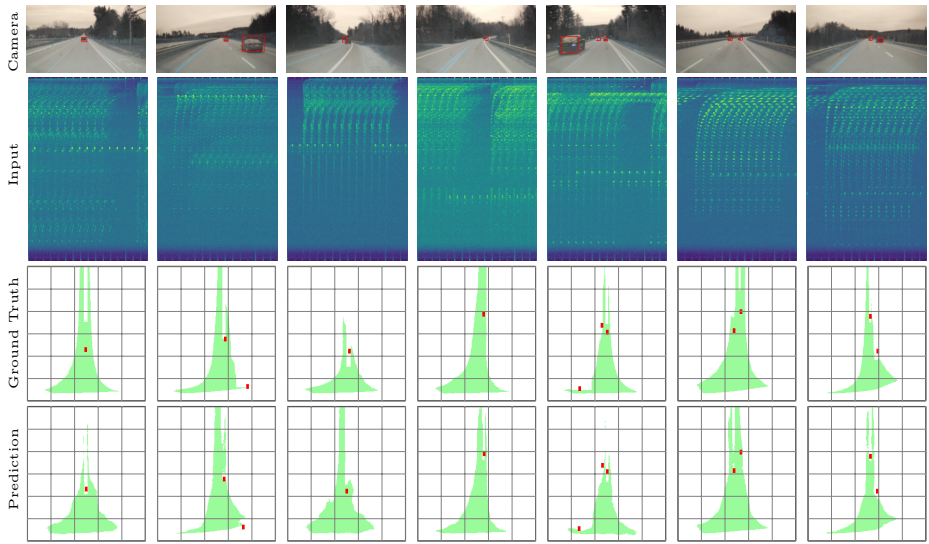
\begin{figure}[tbp] 
    \centering
    \rotatebox{90}{\tiny Camera}
    \includegraphics[width=0.13\textwidth]{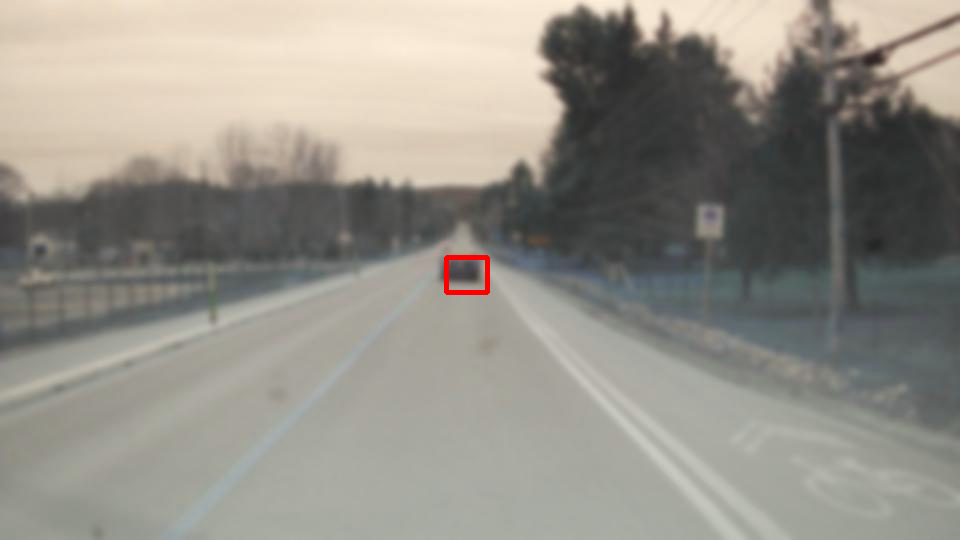}\hspace{1pt}
    \includegraphics[width=0.13\textwidth]{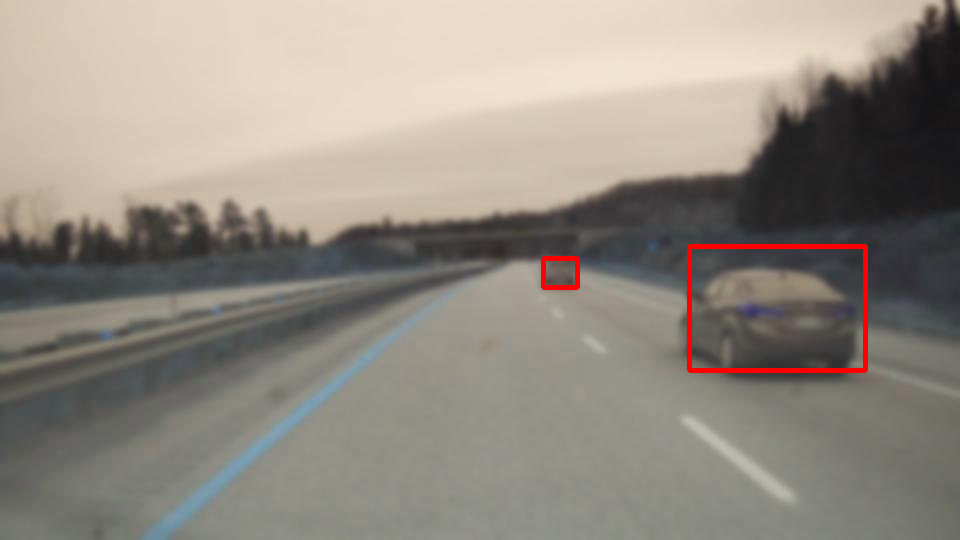}\hspace{1pt}
    \includegraphics[width=0.13\textwidth]{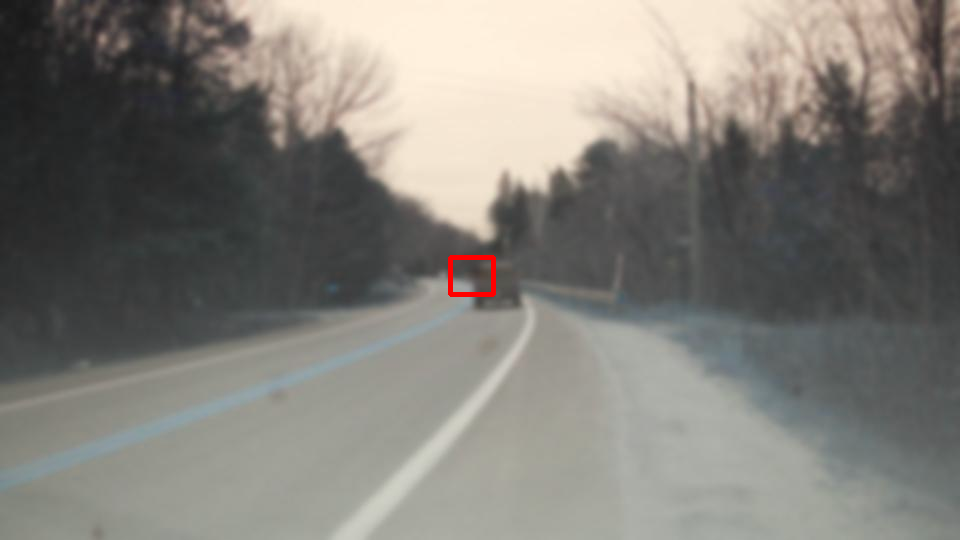}\hspace{1pt}
    \includegraphics[width=0.13\textwidth]{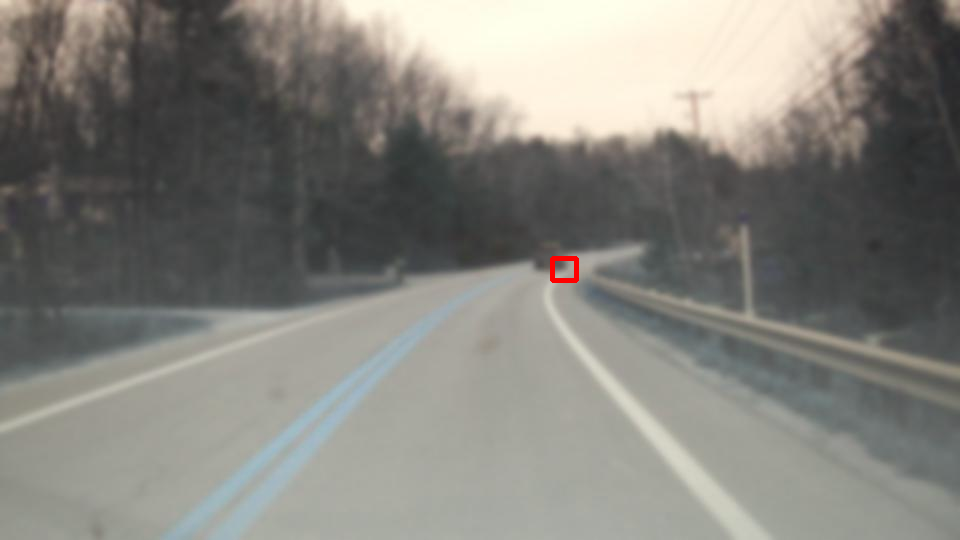}\hspace{1pt}
    \includegraphics[width=0.13\textwidth]{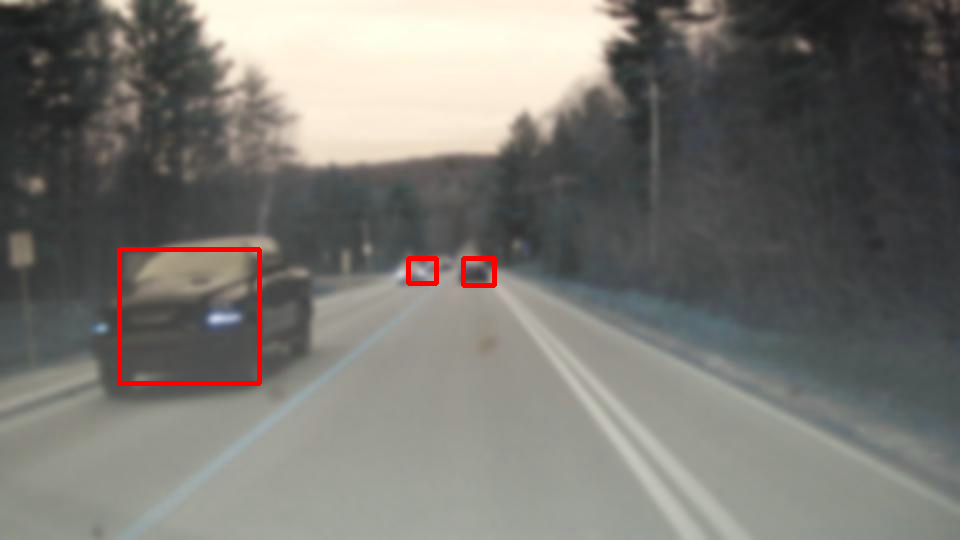}\hspace{1pt}
    \includegraphics[width=0.13\textwidth]{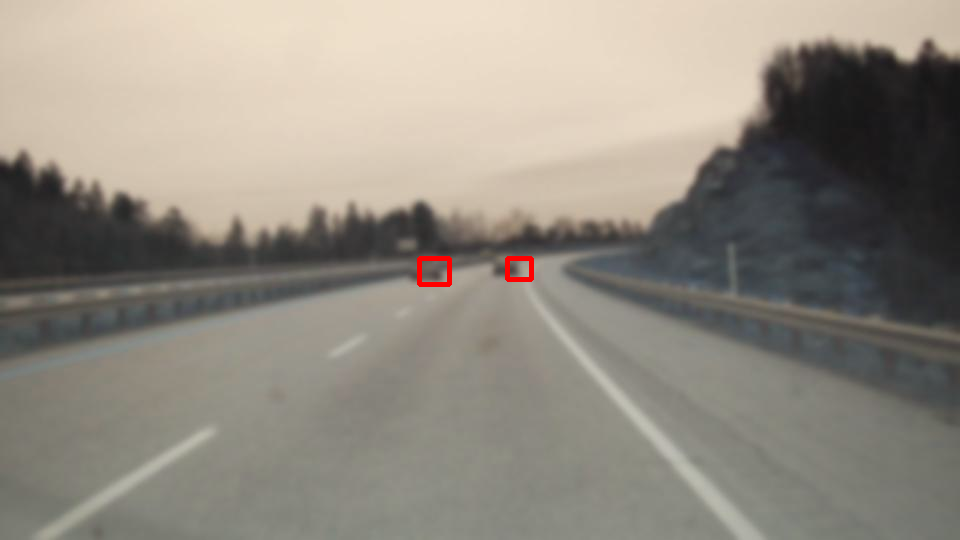}\hspace{1pt}
    \includegraphics[width=0.13\textwidth]{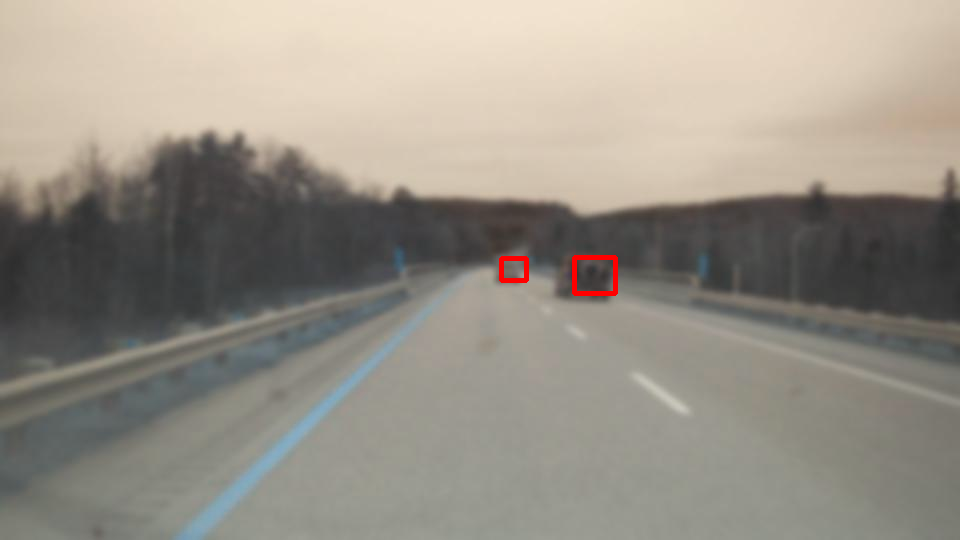}\vspace{1pt}
    \\
    \rotatebox{90}{\tiny ~~~~~~~~~~~~Input}
    \includegraphics[width=0.13\textwidth, height=0.20\textwidth]{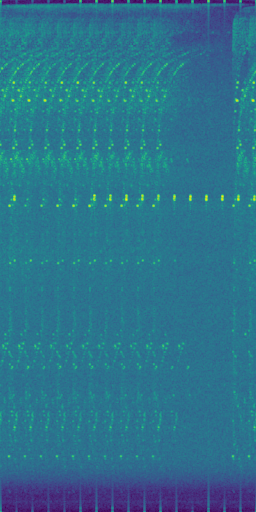}\hspace{1pt}
    \includegraphics[width=0.13\textwidth, height=0.20\textwidth]{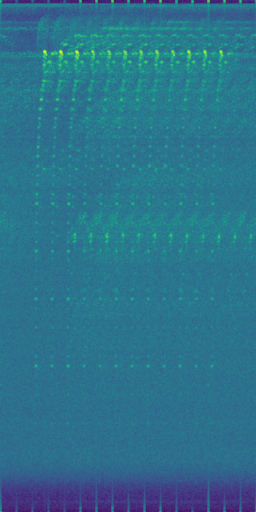}\hspace{1pt}
    \includegraphics[width=0.13\textwidth, height=0.20\textwidth]{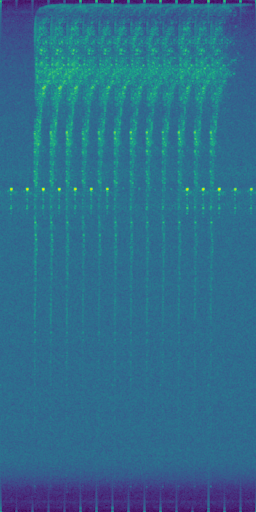}\hspace{1pt}
    \includegraphics[width=0.13\textwidth, height=0.20\textwidth]{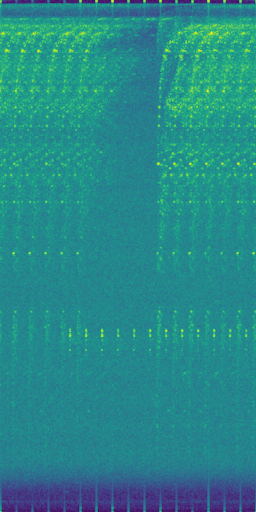}\hspace{1pt}
    \includegraphics[width=0.13\textwidth, height=0.20\textwidth]{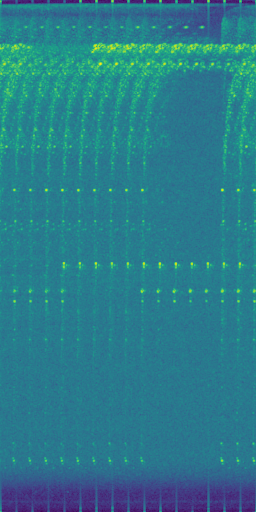}\hspace{1pt}
    \includegraphics[width=0.13\textwidth, height=0.20\textwidth]{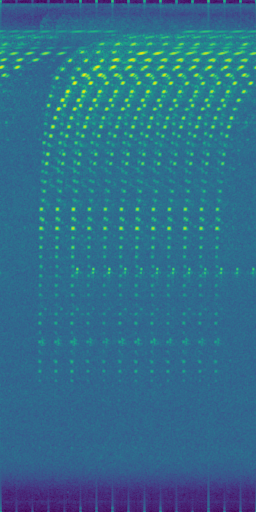}\hspace{1pt}
    \includegraphics[width=0.13\textwidth, height=0.20\textwidth]{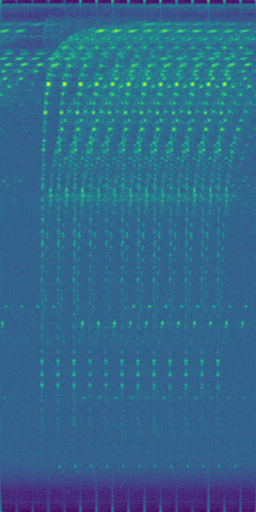}\vspace{1pt}
    \\
    \rotatebox{90}{\tiny ~~Ground Truth}
    \resizebox{0.13\textwidth}{!}{
        \begin{tikzpicture}
            \node[anchor=south west,inner sep=0] (image) at (0,0) {\includegraphics[width=0.9\textwidth,frame]{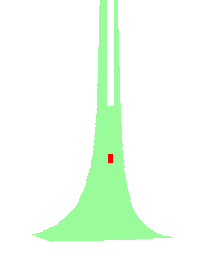}};
            \begin{scope}[x={(image.south east)},y={(image.north west)}]
                \draw[help lines,xstep=.2,ystep=.166] (0,0) grid (1,1);
            \end{scope}
        \end{tikzpicture}}\hspace{1pt}
    \resizebox{0.13\textwidth}{!}{
        \begin{tikzpicture}
            \node[anchor=south west,inner sep=0] (image) at (0,0) {\includegraphics[width=0.9\textwidth,frame]{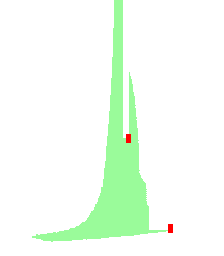}};
            \begin{scope}[x={(image.south east)},y={(image.north west)}]
                \draw[help lines,xstep=.2,ystep=.166] (0,0) grid (1,1);
            \end{scope}
        \end{tikzpicture}}\hspace{1pt}
    \resizebox{0.13\textwidth}{!}{
        \begin{tikzpicture}
            \node[anchor=south west,inner sep=0] (image) at (0,0) {\includegraphics[width=0.9\textwidth,frame]{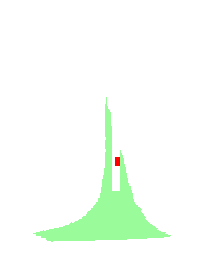}};
            \begin{scope}[x={(image.south east)},y={(image.north west)}]
                \draw[help lines,xstep=.2,ystep=.166] (0,0) grid (1,1);
            \end{scope}
        \end{tikzpicture}}\hspace{1pt}
    \resizebox{0.13\textwidth}{!}{
        \begin{tikzpicture}
            \node[anchor=south west,inner sep=0] (image) at (0,0) {\includegraphics[width=0.9\textwidth,frame]{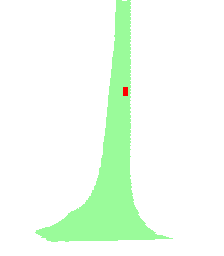}};
            \begin{scope}[x={(image.south east)},y={(image.north west)}]
                \draw[help lines,xstep=.2,ystep=.166] (0,0) grid (1,1);
            \end{scope}
        \end{tikzpicture}}\hspace{1pt}
    \resizebox{0.13\textwidth}{!}{
        \begin{tikzpicture}
            \node[anchor=south west,inner sep=0] (image) at (0,0) {\includegraphics[width=0.9\textwidth,frame]{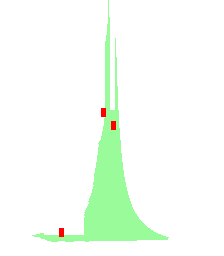}};
            \begin{scope}[x={(image.south east)},y={(image.north west)}]
                \draw[help lines,xstep=.2,ystep=.166] (0,0) grid (1,1);
            \end{scope}
        \end{tikzpicture}}\hspace{1pt}
    \resizebox{0.13\textwidth}{!}{
        \begin{tikzpicture}
            \node[anchor=south west,inner sep=0] (image) at (0,0) {\includegraphics[width=0.9\textwidth,frame]{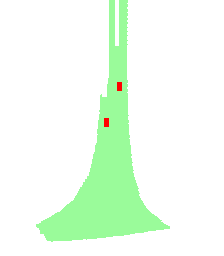}};
            \begin{scope}[x={(image.south east)},y={(image.north west)}]
                \draw[help lines,xstep=.2,ystep=.166] (0,0) grid (1,1);
            \end{scope}
        \end{tikzpicture}}\hspace{1pt}
    \resizebox{0.13\textwidth}{!}{
        \begin{tikzpicture}
            \node[anchor=south west,inner sep=0] (image) at (0,0) {\includegraphics[width=0.9\textwidth,frame]{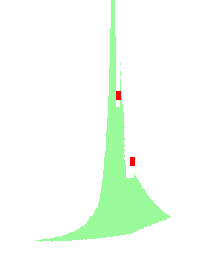}};
            \begin{scope}[x={(image.south east)},y={(image.north west)}]
                \draw[help lines,xstep=.2,ystep=.166] (0,0) grid (1,1);
            \end{scope}
        \end{tikzpicture}}\vspace{1pt}
    \\
    \rotatebox{90}{\tiny ~~~~~Prediction}
    \resizebox{0.13\textwidth}{!}{
        \begin{tikzpicture}
            \node[anchor=south west,inner sep=0] (image) at (0,0) {\includegraphics[width=0.9\textwidth,frame]{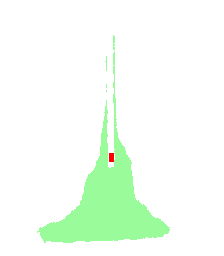}};
            \begin{scope}[x={(image.south east)},y={(image.north west)}]
                \draw[help lines,xstep=.2,ystep=.166] (0,0) grid (1,1);
            \end{scope}
        \end{tikzpicture}}\hspace{1pt}
    \resizebox{0.13\textwidth}{!}{
        \begin{tikzpicture}
            \node[anchor=south west,inner sep=0] (image) at (0,0) {\includegraphics[width=0.9\textwidth,frame]{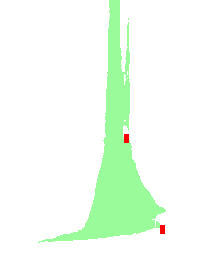}};
            \begin{scope}[x={(image.south east)},y={(image.north west)}]
                \draw[help lines,xstep=.2,ystep=.166] (0,0) grid (1,1);
            \end{scope}
        \end{tikzpicture}}\hspace{1pt}
    \resizebox{0.13\textwidth}{!}{
        \begin{tikzpicture}
            \node[anchor=south west,inner sep=0] (image) at (0,0) {\includegraphics[width=0.9\textwidth,frame]{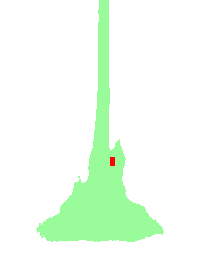}};
            \begin{scope}[x={(image.south east)},y={(image.north west)}]
                \draw[help lines,xstep=.2,ystep=.166] (0,0) grid (1,1);
            \end{scope}
        \end{tikzpicture}}\hspace{1pt}
    \resizebox{0.13\textwidth}{!}{
        \begin{tikzpicture}
            \node[anchor=south west,inner sep=0] (image) at (0,0) {\includegraphics[width=0.9\textwidth,frame]{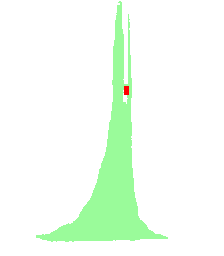}};
            \begin{scope}[x={(image.south east)},y={(image.north west)}]
                \draw[help lines,xstep=.2,ystep=.166] (0,0) grid (1,1);
            \end{scope}
        \end{tikzpicture}}\hspace{1pt}
    \resizebox{0.13\textwidth}{!}{
        \begin{tikzpicture}
            \node[anchor=south west,inner sep=0] (image) at (0,0) {\includegraphics[width=0.9\textwidth,frame]{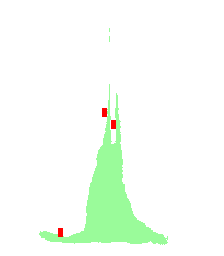}};
            \begin{scope}[x={(image.south east)},y={(image.north west)}]
                \draw[help lines,xstep=.2,ystep=.166] (0,0) grid (1,1);
            \end{scope}
        \end{tikzpicture}}\hspace{1pt}
    \resizebox{0.13\textwidth}{!}{
        \begin{tikzpicture}
            \node[anchor=south west,inner sep=0] (image) at (0,0) {\includegraphics[width=0.9\textwidth,frame]{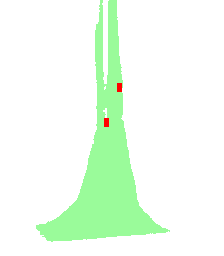}};
            \begin{scope}[x={(image.south east)},y={(image.north west)}]
                \draw[help lines,xstep=.2,ystep=.166] (0,0) grid (1,1);
            \end{scope}
        \end{tikzpicture}}\hspace{1pt}
    \resizebox{0.13\textwidth}{!}{
        \begin{tikzpicture}
            \node[anchor=south west,inner sep=0] (image) at (0,0) {\includegraphics[width=0.9\textwidth,frame]{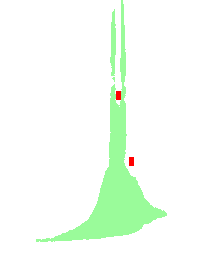}};
            \begin{scope}[x={(image.south east)},y={(image.north west)}]
                \draw[help lines,xstep=.2,ystep=.166] (0,0) grid (1,1);
            \end{scope}
        \end{tikzpicture}}
    \\
    \caption{Qualitative results on the RADIal test set. Camera images are only for display. Predictions are filtered with detection score 0.1 and segmentation score 0.5.} 
    \label{fig:sec4_qualitative}
\end{figure}

Table \ref{sec4:table_semseg} lists the freespace segmentation performances. Since radar raw data is challenging for humans to interpret, the free driving space labels in the RADIal dataset are generated entirely by an image segmentation model and then projected (see Fig. \ref{fig:sec4_qualitative} for examples). Our model uses subsampled input and retains sparsity in the two-branch backbone. As a result, for a dense perception task like segmentation, it exhibits less improvement but still achieves competitive results on the freespace segmentation task.

\subsection{Ablation Study}  \label{sec:4.4} 

We present ablation studies to demonstrate the effectiveness of the DRS module compared with traditional sub-selection methods, and the two-branch backbone compared with different backbone choices.

\textbf{Effect of Deep Radar Subsampling.} We train the DRS module by plugging it into the baseline model FFT-RadNet \cite{FFT-RadNet} and feeding the model with the sparse representation $\bold{y}$. The CA-CFAR algorithm \cite{CACFAR} is a traditional radar data selection approach. It considers a pixel to contain target responses if its signal-to-noise ratio (SNR) is above a certain threshold. Instead of using a fixed SNR threshold, we collect $M$ pixels with the highest SNR to make a fair comparison. Since target reflections of vehicles usually have higher energy, we also include the constant threshold method \cite{constThresh} in the comparison, where we select pixels with top $M$ energy. Similarly, We generate hard sampling masks for these two classical methods, multiplying with input RD spectra to produce sparse representations as input to the baseline model.

\begin{table}[bp]\centering
    \caption{Comparison of radar data sampling methods on the RADIal test set. All methods share the same perception model structure. ``$\dag$'': result reproduced by us.}
    \label{sec4:table_DRS}
    \resizebox{0.5\textwidth}{!}{
    \begin{tabular}{c|c|cccccc}
\toprule
\hline
\#sample             & Method      & F1$\uparrow$ & AP$\uparrow$ & AR$\uparrow$ & RE$\downarrow$ & AE$\downarrow$ & mIoU$\uparrow$  \\ 
 \hline
\multirow{3}{*}{1000} & topM energy & 82.15          & 90.69 & 75.09 & 0.13  & 0.11 & 74.24          \\
                      & CA-CFAR     & 84.62          & 92.67 & 77.85 & 0.13  & 0.12 & 73.35          \\
                      & DRS (ours)  & \textbf{88.13} & 90.99 & 85.44 & 0.15  & 0.10 & \textbf{77.85} \\ \hline
\multirow{3}{*}{4000} & topM energy & 87.93          & 93.78 & 82.76 & 0.12  & 0.11 & 76.62          \\
                      & CA-CFAR     & 86.93          & 94.35 & 80.60  & 0.12  & 0.11 & 75.92          \\
                      & DRS (ours)  & \textbf{90.26} & 93.60  & 87.14 & 0.12  & 0.11 & \textbf{78.05} \\ \hline
512*256               & FFT-RadNet\textsuperscript{\textdagger} \cite{FFT-RadNet} & 90.75          & 95.03 & 86.83 & 0.12  & 0.11 & 77.37          \\ \hline
\bottomrule
    \end{tabular}
    }
\end{table}

\begin{figure}[htbp] \centering
\includegraphics[width=0.5\textwidth]{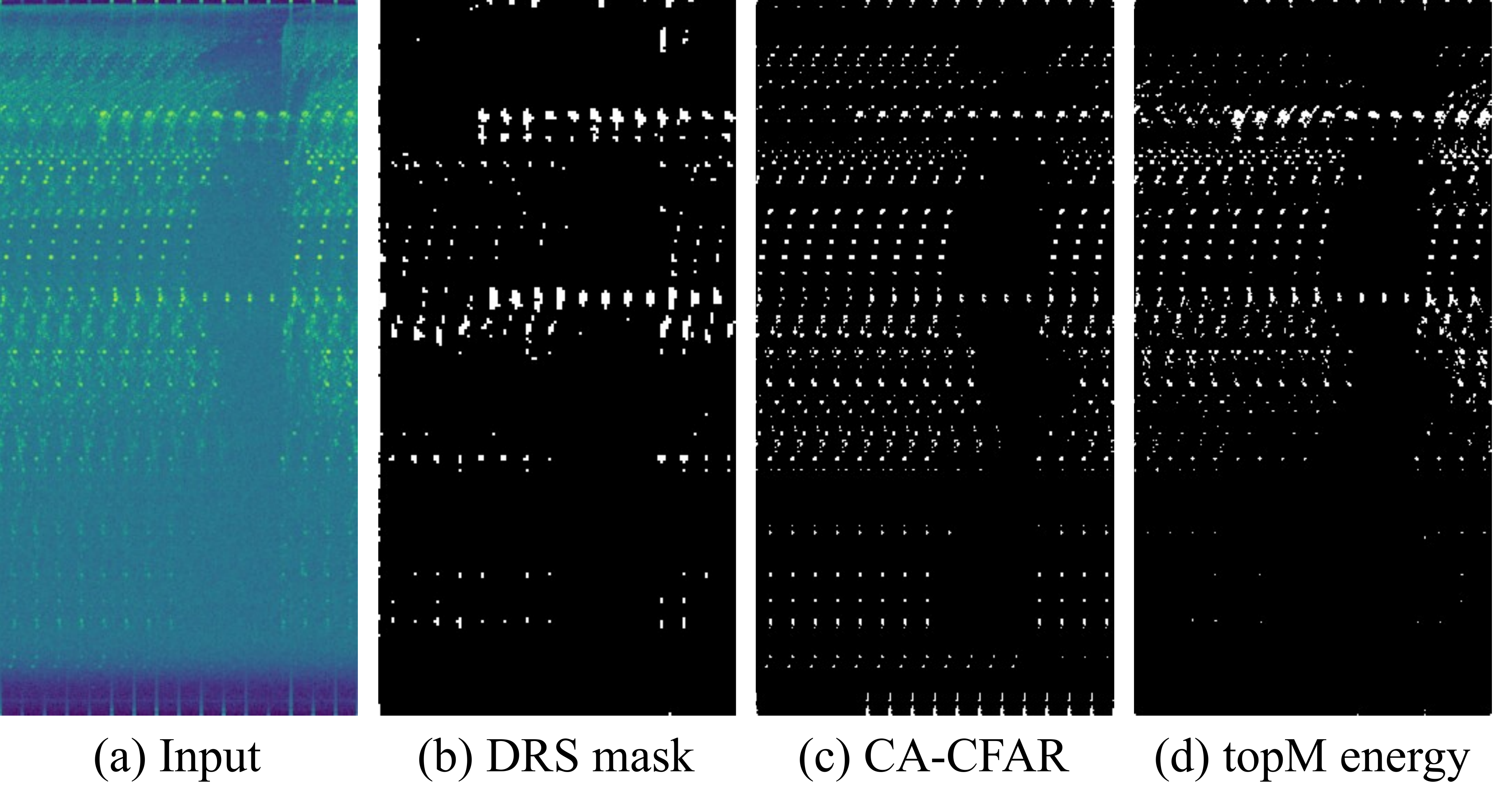}

    \caption{Sampling masks generated by different methods.} 
    \label{fig:sec4_DRS}
\end{figure}

We test with $M=$ 1000 and 4000 for all three subsampling methods, the results are presented in \cref{sec4:table_DRS}. Figure \ref{fig:sec4_DRS} depicts sampling masks with $M=4000$ generated by different methods for the same input frame. Our DRS approach selects a task-aware subset of pixels, provides more relevant context for perception, and leads to higher performance. Our result with 4000 samples shows comparable performance to the baseline that takes full RD spectra as input. That indicates the subsampled data carries sufficient information for perception.

\textbf{Effect of the Two-branch Backbone.} The effectiveness of our two-branch backbone is demonstrated in \cref{sec4:table_ablation}. The SCNN-only and GNN-only models are scaled up accordingly to ensure a fair comparison. All methods use subsampled inputs from our DRS module with $M$ set to 4000.

Applying a dense backbone (with DRS) yields a higher segmentation score than sparse SCNN-only and GNN-only models. By replacing CNN layers with SCNN layers, the SCNN-only model requires less computational overhead. However, this reduction comes at the cost of decreased accuracy due to the smaller receptive field of submanifold SCNNs. The GNN-only model improves accuracy by aggregating features of embedding space neighbors. By employing the two-branch backbone, the performance is further enhanced as it incorporates both global embedding space neighbor and local spatial neighbor information. This model uses only the spatial attentive addition in the fusion. Finally, the axial attention fusion module aligns local and global features, maximally leveraging features from different aspects to boost performance to its greatest extent.

\begin{table}[tbp]\centering
    \caption{Ablation study on different backbones on the RADIal test set. DRS means the DRS module plugged into the baseline. AAF denotes the axial attentive fusion. }
    \label{sec4:table_ablation}
    \resizebox{0.5\textwidth}{!}{
\begin{tabular}{cccccccccc}
\toprule
\hline

DRS & SCNN & GNN & AAF & F1$\uparrow$   & AP$\uparrow$& AR$\uparrow$& RE$\downarrow$& AE$\downarrow$& mIoU$\uparrow$ \\ \hline
\checkmark   &      &     &     & 90.26          & 93.60 & 87.14 & 0.12 & 0.11 & 78.05          \\
\checkmark   & \checkmark    &     &     & 89.50          & 93.81 & 85.57 & 0.13 & 0.11 & 76.52          \\
\checkmark   &      & \checkmark   &     & 90.70          & 93.76 & 87.84 & 0.13 & 0.10 & 77.05          \\
\checkmark   & \checkmark    & \checkmark   &     & 91.71          & 95.43 & 88.26 & 0.13 & 0.10 & 78.02          \\
\checkmark   & \checkmark    & \checkmark   & \checkmark   & \textbf{93.84} & 96.00 & 91.78 & 0.13 & 0.10 & \textbf{78.48} \\ \hline
    \bottomrule
\end{tabular}%
    }
\end{table}

\subsection{Complexity analysis} \label{sec:4.5}

Table \ref{sec4:table_flops} compares FLOPs (Floating Point Operations) and the number of parameters of models. Since FLOPs of SCNNs depend on the input, we calculate it by averaging over the test set. Compared to the baseline FFT-RadNet \cite{FFT-RadNet}, our SparseRadNet introduces additional modules including the DRS module, GNN blocks, attentive fusion, and our backbone has two branches. These modules result in an increase in network parameters to 6.9M. Nevertheless, by harnessing the sparsity of radar data, our model achieves significantly higher accuracy and requires 10\% less computational cost compared to FFT-RadNet.

\begin{table}[tbp]\centering
    \caption{Complexity analysis in terms of FLOPs and number of parameters. ``$\dag$'': result reproduced by us. }
    \label{sec4:table_flops}
    \resizebox{0.35\textwidth}{!}{
\begin{tabular}{ccc}
\toprule
\hline
Model               & FLOPs$\downarrow$ & Params$\downarrow$ \\ \hline
FFT-RadNet \cite{FFT-RadNet}     & 288G  & 3.8M  \\
T-FFTRadNet \cite{T-FFTRadNet}& \textbf{194G}  & 9.6M  \\
Cross Modal DNN \cite{XModSup}& 358G  & 7.7M  \\
TransRadar\textsuperscript{\textdagger} \cite{TransRadar} & 343G  & 3.7M  \\
SparseRadNet (ours)  & \underline{259G}  & 6.9M  \\

\hline
\bottomrule
\end{tabular}%
    }
\end{table}

\section{Conclusion}%
\label{sec:Conclusion}
We introduced SparseRadNet, a novel radar perception model. Our model exploits the sparse nature of radar data, identifying crucial components from sparse signals and exploring different levels of affinity using a specifically tailored network architecture. Our deep radar subsampling module selects a subset of pixels that contribute most to the perception task, and our two-branch backbone captures local and global dependencies. Those are fused using an attentive fusion module. Experiments conducted on the RADIal dataset show that SparseRadNet outperforms previous models in object detection. Ablation studies demonstrate that our DRS module, sampling only 3\% of input, yet provides sufficient information for perception, and our two-branch backbone enhances accuracy by aggregating local and global neighbor information.

\section*{Acknowledgements}
J.W.\ and M.R.\ acknowledge support by the German Federal Ministry of Education and Research within the junior research group project “UnrEAL” (grant no.\ 01IS22069).

%
%
\bibliographystyle{splncs04}
\bibliography{main}

\begin{thebibliography}{10}
\providecommand{\url}[1]{\texttt{#1}}
\providecommand{\urlprefix}{URL }
\providecommand{\doi}[1]{https://doi.org/#1}

\bibitem{Pointillism}
Bansal, K., Rungta, K., Zhu, S., Bharadia, D.: Pointillism: Accurate 3d bounding box estimation with multi-radars. In: Proceedings of the 18th Conference on Embedded Networked Sensor Systems. pp. 340--353 (2020)

\bibitem{FMCW}
Brooker, G.M., et~al.: Understanding millimetre wave fmcw radars. In: 1st international Conference on Sensing Technology. vol.~1 (2005)

\bibitem{ASPP}
Chen, L.C., Papandreou, G., Kokkinos, I., Murphy, K., Yuille, A.L.: Deeplab: Semantic image segmentation with deep convolutional nets, atrous convolution, and fully connected crfs. IEEE transactions on pattern analysis and machine intelligence  \textbf{40}(4),  834--848 (2017)

\bibitem{spconv}
Contributors, S.: Spconv: Spatially sparse convolution library. \url{https://github.com/traveller59/spconv} (2022)

\bibitem{TransRadar}
Dalbah, Y., Lahoud, J., Cholakkal, H.: Transradar: Adaptive-directional transformer for real-time multi-view radar semantic segmentation. In: Proceedings of the IEEE/CVF Winter Conference on Applications of Computer Vision. pp. 353--362 (2024)

\bibitem{PointNetIDP}
Danzer, A., Griebel, T., Bach, M., Dietmayer, K.: 2d car detection in radar data with pointnets. In: 2019 IEEE Intelligent Transportation Systems Conference (ITSC). pp. 61--66. IEEE (2019)

\bibitem{DAROD}
Decourt, C., VanRullen, R., Salle, D., Oberlin, T.: Darod: A deep automotive radar object detector on range-doppler maps. In: 2022 IEEE Intelligent Vehicles Symposium (IV). pp. 112--118. IEEE (2022)

\bibitem{RadarUncer}
Dong, X., Wang, P., Zhang, P., Liu, L.: Probabilistic oriented object detection in automotive radar. In: Proceedings of the IEEE/CVF Conference on Computer Vision and Pattern Recognition Workshops. pp. 102--103 (2020)

\bibitem{2DProject}
Dreher, M., Er{\c{c}}elik, E., B{\"a}nziger, T., Knoll, A.: Radar-based 2d car detection using deep neural networks. In: 2020 IEEE 23rd International Conference on Intelligent Transportation Systems (ITSC). pp.~1--8. IEEE (2020)

\bibitem{RadarGNN}
Fent, F., Bauerschmidt, P., Lienkamp, M.: Radargnn: Transformation invariant graph neural network for radar-based perception. In: Proceedings of the IEEE/CVF Conference on Computer Vision and Pattern Recognition. pp. 182--191 (2023)

\bibitem{RAMP-CNN}
Gao, X., Xing, G., Roy, S., Liu, H.: Ramp-cnn: A novel neural network for enhanced automotive radar object recognition. IEEE Sensors Journal  \textbf{21}(4),  5119--5132 (2020)

\bibitem{T-FFTRadNet}
Giroux, J., Bouchard, M., Laganiere, R.: T-fftradnet: Object detection with swin vision transformers from raw adc radar signals. In: Proceedings of the IEEE/CVF International Conference on Computer Vision. pp. 4030--4039 (2023)

\bibitem{submanifold}
Graham, B., Van~der Maaten, L.: Submanifold sparse convolutional networks. arXiv preprint arXiv:1706.01307  (2017)

\bibitem{GumbelTrick}
Gumbel, E.J.: Statistical theory of extreme values and some practical applications: a series of lectures, vol.~33. US Government Printing Office (1948)

\bibitem{ViG}
Han, K., Wang, Y., Guo, J., Tang, Y., Wu, E.: Vision gnn: An image is worth graph of nodes. Advances in Neural Information Processing Systems  \textbf{35},  8291--8303 (2022)

\bibitem{ResNet}
He, K., Zhang, X., Ren, S., Sun, J.: Deep residual learning for image recognition. In: Proceedings of the IEEE conference on computer vision and pattern recognition. pp. 770--778 (2016)

\bibitem{SVGA-Net}
He, Q., Wang, Z., Zeng, H., Zeng, Y., Liu, Y.: Svga-net: Sparse voxel-graph attention network for 3d object detection from point clouds. In: Proceedings of the AAAI Conference on Artificial Intelligence. vol.~36, pp. 870--878 (2022)

\bibitem{GeLU}
Hendrycks, D., Gimpel, K.: Gaussian error linear units (gelus). arXiv preprint arXiv:1606.08415  (2016)

\bibitem{AxAtt}
Ho, J., Kalchbrenner, N., Weissenborn, D., Salimans, T.: Axial attention in multidimensional transformers. arXiv preprint arXiv:1912.12180  (2019)

\bibitem{DPS}
Huijben, I.A., Veeling, B.S., van Sloun, R.J.: Deep probabilistic subsampling for task-adaptive compressed sensing. In: International Conference on Learning Representations (2019)

\bibitem{CACFAR}
Jalil, A., Yousaf, H., Baig, M.I.: Analysis of cfar techniques. In: 2016 13th International Bhurban Conference on Applied Sciences and Technology (IBCAST). pp. 654--659. IEEE (2016)

\bibitem{GumbelSoftmax}
Jang, E., Gu, S., Poole, B.: Categorical reparameterization with gumbel-softmax. In: International Conference on Learning Representations (2017)

\bibitem{XModSup}
Jin, Y., Deligiannis, A., Fuentes-Michel, J.C., Vossiek, M.: Cross-modal supervision-based multitask learning with automotive radar raw data. IEEE Transactions on Intelligent Vehicles  (2023)

\bibitem{constThresh}
Jose, E., Adams, M.D.: Millimetre wave radar spectra simulation and interpretation for outdoor slam. In: IEEE International Conference on Robotics and Automation, 2004. Proceedings. ICRA'04. 2004. vol.~2, pp. 1321--1326. IEEE (2004)

\bibitem{TRL}
Li, P., Wang, P., Berntorp, K., Liu, H.: Exploiting temporal relations on radar perception for autonomous driving. In: Proceedings of the IEEE/CVF Conference on Computer Vision and Pattern Recognition. pp. 17071--17080 (2022)

\bibitem{Swin}
Liu, Z., Lin, Y., Cao, Y., Hu, H., Wei, Y., Zhang, Z., Lin, S., Guo, B.: Swin transformer: Hierarchical vision transformer using shifted windows. In: Proceedings of the IEEE/CVF international conference on computer vision. pp. 10012--10022 (2021)

\bibitem{Radatron}
Madani, S., Guan, J., Ahmed, W., Gupta, S., Hassanieh, H.: Radatron: Accurate detection using multi-resolution cascaded mimo radar. In: European Conference on Computer Vision. pp. 160--178. Springer (2022)

\bibitem{3ViewConcat}
Major, B., Fontijne, D., Ansari, A., Teja~Sukhavasi, R., Gowaikar, R., Hamilton, M., Lee, S., Grzechnik, S., Subramanian, S.: Vehicle detection with automotive radar using deep learning on range-azimuth-doppler tensors. In: Proceedings of the IEEE/CVF International Conference on Computer Vision Workshops. pp.~0--0 (2019)

\bibitem{RadGCN}
Meyer, M., Kuschk, G., Tomforde, S.: Graph convolutional networks for 3d object detection on radar data. In: Proceedings of the IEEE/CVF International Conference on Computer Vision. pp. 3060--3069 (2021)

\bibitem{MobileViG}
Munir, M., Avery, W., Marculescu, R.: Mobilevig: Graph-based sparse attention for mobile vision applications. In: Proceedings of the IEEE/CVF Conference on Computer Vision and Pattern Recognition. pp. 2210--2218 (2023)

\bibitem{MVRSS}
Ouaknine, A., Newson, A., P{\'e}rez, P., Tupin, F., Rebut, J.: Multi-view radar semantic segmentation. In: Proceedings of the IEEE/CVF International Conference on Computer Vision. pp. 15671--15680 (2021)

\bibitem{carrada}
Ouaknine, A., Newson, A., Rebut, J., Tupin, F., P{\'e}rez, P.: Carrada dataset: Camera and automotive radar with range-angle-doppler annotations. In: 2020 25th International Conference on Pattern Recognition (ICPR). pp. 5068--5075. IEEE (2021)

\bibitem{RTCNet}
Palffy, A., Dong, J., Kooij, J.F., Gavrila, D.M.: Cnn based road user detection using the 3d radar cube. IEEE Robotics and Automation Letters  \textbf{5}(2),  1263--1270 (2020)

\bibitem{RadarPillars}
Palffy, A., Pool, E., Baratam, S., Kooij, J.F., Gavrila, D.M.: Multi-class road user detection with 3+ 1d radar in the view-of-delft dataset. IEEE Robotics and Automation Letters  \textbf{7}(2),  4961--4968 (2022)

\bibitem{PointNet++}
Qi, C.R., Yi, L., Su, H., Guibas, L.J.: Pointnet++: Deep hierarchical feature learning on point sets in a metric space. Advances in neural information processing systems  \textbf{30} (2017)

\bibitem{FFT-RadNet}
Rebut, J., Ouaknine, A., Malik, W., P{\'e}rez, P.: Raw high-definition radar for multi-task learning. In: Proceedings of the IEEE/CVF Conference on Computer Vision and Pattern Recognition. pp. 17021--17030 (2022)

\bibitem{FasterR-CNN}
Ren, S., He, K., Girshick, R., Sun, J.: Faster r-cnn: Towards real-time object detection with region proposal networks. Advances in neural information processing systems  \textbf{28} (2015)

\bibitem{mimoRef2}
Robey, F.C., Coutts, S., Weikle, D., McHarg, J.C., Cuomo, K.: Mimo radar theory and experimental results. In: Conference Record of the Thirty-Eighth Asilomar Conference on Signals, Systems and Computers, 2004. vol.~1, pp. 300--304. IEEE (2004)

\bibitem{UNet}
Ronneberger, O., Fischer, P., Brox, T.: U-net: Convolutional networks for biomedical image segmentation. In: Medical Image Computing and Computer-Assisted Intervention--MICCAI 2015: 18th International Conference, Munich, Germany, October 5-9, 2015, Proceedings, Part III 18. pp. 234--241. Springer (2015)

\bibitem{DynStatic}
Schumann, O., Lombacher, J., Hahn, M., W{\"o}hler, C., Dickmann, J.: Scene understanding with automotive radar. IEEE Transactions on Intelligent Vehicles  \textbf{5}(2),  188--203 (2019)

\bibitem{RADIATEdataset}
Sheeny, M., De~Pellegrin, E., Mukherjee, S., Ahrabian, A., Wang, S., Wallace, A.: Radiate: A radar dataset for automotive perception in bad weather. In: 2021 IEEE International Conference on Robotics and Automation (ICRA). pp.~1--7. IEEE (2021)

\bibitem{Point-GNN}
Shi, W., Rajkumar, R.: Point-gnn: Graph neural network for 3d object detection in a point cloud. In: Proceedings of the IEEE/CVF conference on computer vision and pattern recognition. pp. 1711--1719 (2020)

\bibitem{RadarMFNet}
Tan, B., Ma, Z., Zhu, X., Li, S., Zheng, L., Chen, S., Huang, L., Bai, J.: 3d object detection for multi-frame 4d automotive millimeter-wave radar point cloud. IEEE Sensors Journal  (2022)

\bibitem{KPConv}
Thomas, H., Qi, C.R., Deschaud, J.E., Marcotegui, B., Goulette, F., Guibas, L.J.: Kpconv: Flexible and deformable convolution for point clouds. In: Proceedings of the IEEE/CVF international conference on computer vision. pp. 6411--6420 (2019)

\bibitem{KPConvPillars}
Ulrich, M., Braun, S., K{\"o}hler, D., Niederl{\"o}hner, D., Faion, F., Gl{\"a}ser, C., Blume, H.: Improved orientation estimation and detection with hybrid object detection networks for automotive radar. In: 2022 IEEE 25th International Conference on Intelligent Transportation Systems (ITSC). pp. 111--117. IEEE (2022)

\bibitem{ADPS}
Van~Gorp, H., Huijben, I., Veeling, B.S., Pezzotti, N., Van~Sloun, R.J.: Active deep probabilistic subsampling. In: International Conference on Machine Learning. pp. 10509--10518. PMLR (2021)

\bibitem{RODNet}
Wang, Y., Jiang, Z., Li, Y., Hwang, J.N., Xing, G., Liu, H.: Rodnet: A real-time radar object detection network cross-supervised by camera-radar fused object 3d localization. IEEE Journal of Selected Topics in Signal Processing  \textbf{15}(4),  954--967 (2021)

\bibitem{mmwaveReview}
Wei, Z., Zhang, F., Chang, S., Liu, Y., Wu, H., Feng, Z.: Mmwave radar and vision fusion for object detection in autonomous driving: A review. Sensors  \textbf{22}(7), ~2542 (2022)

\bibitem{SpatialAtt}
Woo, S., Park, J., Lee, J.Y., Kweon, I.S.: Cbam: Convolutional block attention module. In: Proceedings of the European Conference on Computer Vision (ECCV) (September 2018)

\bibitem{groupedConv}
Xie, S., Girshick, R., Doll{\'a}r, P., Tu, Z., He, K.: Aggregated residual transformations for deep neural networks. In: Proceedings of the IEEE conference on computer vision and pattern recognition. pp. 1492--1500 (2017)

\bibitem{ADCNet}
Yang, B., Khatri, I., Happold, M., Chen, C.: Adcnet: Learning from raw radar data via distillation. arXiv preprint arXiv:2303.11420  (2023)

\bibitem{RADDet}
Zhang, A., Nowruzi, F.E., Laganiere, R.: Raddet: Range-azimuth-doppler based radar object detection for dynamic road users. In: 2021 18th Conference on Robots and Vision (CRV). pp. 95--102. IEEE (2021)

\bibitem{PhaseNorm}
Zhang, G., Li, H., Wenger, F.: Object detection and 3d estimation via an fmcw radar using a fully convolutional network. In: ICASSP 2020-2020 IEEE International Conference on Acoustics, Speech and Signal Processing (ICASSP). pp. 4487--4491. IEEE (2020)

\bibitem{PeakConv}
Zhang, L., Zhang, X., Zhang, Y., Guo, Y., Chen, Y., Huang, X., Ma, Z.: Peakconv: Learning peak receptive field for radar semantic segmentation. In: Proceedings of the IEEE/CVF Conference on Computer Vision and Pattern Recognition. pp. 17577--17586 (2023)

\bibitem{review}
Zhou, Y., Liu, L., Zhao, H., L{\'o}pez-Ben{\'\i}tez, M., Yu, L., Yue, Y.: Towards deep radar perception for autonomous driving: Datasets, methods, and challenges. Sensors  \textbf{22}(11), ~4208 (2022)

\end{thebibliography}

\clearpage
\setcounter{page}{1}

\begin{center}
\large
\textbf{\large SparseRadNet: Sparse Perception Neural Network on Subsampled Radar Data} \\
        \vspace{0.5em}Supplementary Material \\
        \vspace{1.0em}
\end{center}

\appendix

\section{More Details of Experiments} \label{appendix_A}

\subsection{Results of Reproduced Baseline Models}

\subsubsection{FFT-RadNet.} Even when random seeds are set, training radar perception models exhibits higher variance than LiDAR and camera models. We reproduced the training of the baseline model FFT-RadNet and achieved a higher F1 score of 90.75 compared to the reported results of 88.91 \cite{FFT-RadNet}. When pre-training our deep radar subsampling (DRS) module using the FFT-RadNet baseline, we also achieved a higher F1 score of 90.26 compared to the reported FFT-RadNet. However, the accuracy is lower than our reproduced FFT-RadNet due to the subsampling of input (see \cref{append:fftradnet}). 

\begin{table}[htbp]\centering
    \caption{Reproduction of FFT-RadNet \cite{FFT-RadNet}. ``$\dag$'': result reproduced by us.  The DRS module samples M=4000 pixels as the sparse input.}
    \label{append:fftradnet}
    \resizebox{0.55\textwidth}{!}{
\begin{tabular}{c|cccccc}
\hline
Method                  & F1$\uparrow$ & AP$\uparrow$ & AR$\uparrow$ & RE$\downarrow$ & AE$\downarrow$ & mIoU$\uparrow$           \\ \hline
FFT-RadNet \cite{FFT-RadNet} & 88.91          & 96.84 & 82.18 & 0.11 & 0.17 & 74.00          \\ \hline
DRS + FFT-RadNet        & 90.26          & 93.60 & 87.14 & 0.12 & 0.11 & \textbf{78.05} \\
FFT-RadNet\textsuperscript{\textdagger}   & \textbf{90.75} & 95.03 & 86.83 & 0.12 & 0.11 & 77.37          \\ \hline
\end{tabular}
    }
\end{table}

\begin{table}[bp]\centering
    \caption{Reproduction of TransRadar\cite{TransRadar}. ``$\dag$'': result reproduced by us. }
    \label{append:transradar}
    \resizebox{0.9\textwidth}{!}{
\begin{tabular}{c|cccccc|cc}
\hline
Method                                   & F1$\uparrow$ & AP$\uparrow$ & AR$\uparrow$ & RE$\downarrow$ & AE$\downarrow$ & mIoU$\uparrow$           & FLOPs$\downarrow$ & Params$\downarrow$ \\ \hline
TransRadar \cite{TransRadar}   & 97.85          & 97.30 & 98.40 & 0.11 & 0.10 & 81.10         & N/A  & 3.4M   \\ \hline
TransRadar\textsuperscript{\textdagger} - {[}128, 64{]} no padding    & 91.95          & 95.00 & 89.09 & 0.15 & 0.11 & 82.20          & \textbf{226G}  & \textbf{3.4M}   \\
TransRadar\textsuperscript{\textdagger} - {[}128, 256{]} no padding   & 91.95          & 94.84 & 89.23 & 0.15 & 0.10 & \textbf{82.57}          & 362G  & 4.3M   \\

TransRadar\textsuperscript{\textdagger} - {[}128, 256{]} downsampling & \textbf{92.86} & 94.91 & 90.90 & 0.15 & 0.10 & 82.27 & 343G  & 3.7M   \\ \hline
\end{tabular}
    }
\end{table}

\subsubsection{TransRadar.} TransRadar \cite{TransRadar} is the previous best model on the RADIal dataset. The numerical results from \cite{TransRadar} as well as the corresponding repository focus on the CARRADA dataset \cite{carrada}. Therefore, we reproduced TransRadar results for the RADIal dataset based on the description from \cite{TransRadar}. In the RADIal dataset, the input range-Doppler (RD) spectra have a spatial shape of $512 \times 256$, with $256$ bins in the Doppler axis; while the required range-angle (RA) space output shapes are $128 \times 224$ for object detection labels and $256 \times 224$ for freespace segmentation labels, with $224$ bins in the angle axis. TransRadar uses an autoencoder structure and does not employ the range-angle decoder \cite{FFT-RadNet} to transform from RD to RA space. Therefore, we experimented with different approaches to match the dimensions of the Doppler ($256$) and angle axis ($224$) in our reproduction. Results on the overall RADIal test set are shown in \cref{append:transradar}. The first reproduced model's spatial size in the latent space is $128 \times 64$, and the dimensions are gradually shrunk by removing paddings of CNN layers in the Doppler axis. The second model's latent space has a spatial size of $128 \times 256$, and dimensions are also matched by zero-padding. The third model achieves the highest object detection accuracy in term of F1 score, and the dimension matching is done by applying downsampling. All three models achieve higher freespace segmentation scores but lower object detection scores compared to the results reported in \cite{TransRadar}.

\subsection{Loss Function}

We use the same loss functions as FFT-RadNet \cite{FFT-RadNet}. For training the object detection head, the Focal loss is used for classification and the smooth L1 loss for range and angle offset regression. For the freespace segmentation head, the binary cross entropy loss is applied. The overall loss is a sum of detection and segmentation losses, which enables simultaneous training of the multi-task head.

\subsection{Evaluation Metric}

We follow the evaluation metrics of the RADIal dataset \cite{FFT-RadNet}. Predictions are firstly filtered based on thresholds, retaining those with detection scores exceeding $0.1$ and freespace segmentation scores exceeding $0.5$. Then the non-maximum suppression (NMS) is applied to remove duplicate predictions. A prediction is considered a true positive if it has an IoU above 50\% with any ground truth bounding box. The average precision (AP) and average recall (AR) are computed by calculating the precision and recall of each frame and then taking an average over all frames, respectively. The average F1 score is directly computed from AP and AR:

\begin{equation}
    F1 = \frac{2 \times AP \times AR}{AP+AR}.
\end{equation}
The IoU between a predicted segmentation map and a ground truth segmentation map is defined as the intersection of the occupied area over the union of the occupied area in the two maps. The mIoU score for freespace segmentation is computed by averaging the IoU over all frames.

\subsection{Additional Ablation Studies}

In Sec. 4.3, the two models with SCNN-only and GNN-only backbones are scaled up accordingly to ensure a fair comparison with the two-branch backbone. The SCNN-only model is scaled up to include 3 and 4 sparse residual blocks in the first two RNBs, using a similar combination of regular and submanifold SCNN layers. The number of GNN blocks in the GNN-only model is increased to eight.

\subsubsection{Impact of the MIMO pre-encoder.}

In order to acquire full gradient information, our DRS module relies on pre-training using a dense backbone. Additionally, the MIMO pre-encoder connects the gradients of repeated object signatures, which are provided to the DRS module during backpropagation. This enables the DRS module to be aware of the repetition in the sampling. Otherwise, the receptive field of the CNN layers in the DRS module is not large enough to cover all repetitions in one row, leading to information loss. To demonstrate the impact of the MIMO pre-encoder in pre-training, we test the pre-training using a dense model without the MIMO pre-encoder, results presented in \cref{append:MIMO}. Without the MIMO-pre-encoder in pre-training, the DRS module fails to provide sufficient information for perception. However, with the inclusion of the MIMO pre-encoder, the model equipped with the DRS module achieves comparable accuracy to the baseline model that uses the full input.

\begin{table}[htbp]\centering
    \caption{Impact of the MIMO pre-encoder on pre-training the DRS module. The DRS module samples M=4000 pixels as the sparse input.}
    \label{append:MIMO}
    \resizebox{0.55\textwidth}{!}{
\begin{tabular}{c|cccccc}
\hline
Method              & F1$\uparrow$ & AP$\uparrow$ & AR$\uparrow$ & RE$\downarrow$ & AE$\downarrow$ & mIoU$\uparrow$  \\ \hline
DRS + dense backbone & 73.94 & 76.81 & 71.28 & 0.21 & 0.13 & 78.09 \\
dense backbone & 90.34 & 93.85 & 87.08 & 0.14 & 0.12 & 74.94  \\
\hline
DRS + FFT-RadNet & 90.26 & 93.60 & 87.14 & 0.12 & 0.11 & 78.05 \\
FFT-RadNet\textsuperscript{\textdagger}   & 90.75 & 95.03 & 86.83 & 0.12 & 0.11 & 77.37 \\
\hline
\end{tabular}
    }

\end{table}

\subsubsection{DRS module.}

Table \ref{append:rebuttal_DRS} shows ablation studies on the DRS module in terms of the temperature parameter $\tau$ and the number of samples $M$. Increasing $\tau$ appropriately smooths the softmax in Eq. (2), increases randomness during training, and enables exploration of a better subset of pixels. Increasing the sample size provides perception with more information.

\begin{table}[htbp]\centering
    \caption{Ablation studies on the DRS module.}
    \label{append:rebuttal_DRS}
    \resizebox{0.63\textwidth}{!}{
\begin{tabular}{c|ccccc|ccccc}
\hline
\multirow{2}{*}{Metric} & \multicolumn{5}{c|}{softmax temperature $\tau$} & \multicolumn{5}{c}{\#samples $M$ in DRS}     \\ \cline{2-11} 
                        & 2       & 4       & 6       & 8       & 10      & 1000  & 3000  & 4000  & 6000  & 8000  \\ \hline
F1$\uparrow$            & 75.08   & \underline{90.26}   & 90.79   & 90.13   & 89.86   & 88.13 & 89.95 & \underline{90.26} & 90.64 & 90.82 \\ \hline
mIoU$\uparrow$          & 77.11   & \underline{78.05}   & 76.79   & 77.47   & 77.15   & 77.85 & 77.04 & \underline{78.05} & 74.06 & 78.33 \\ \hline
\end{tabular}
    }
\end{table}


\subsubsection{Positional encoding.}
\begin{figure}[tbp] \centering
\includegraphics[width=\textwidth]{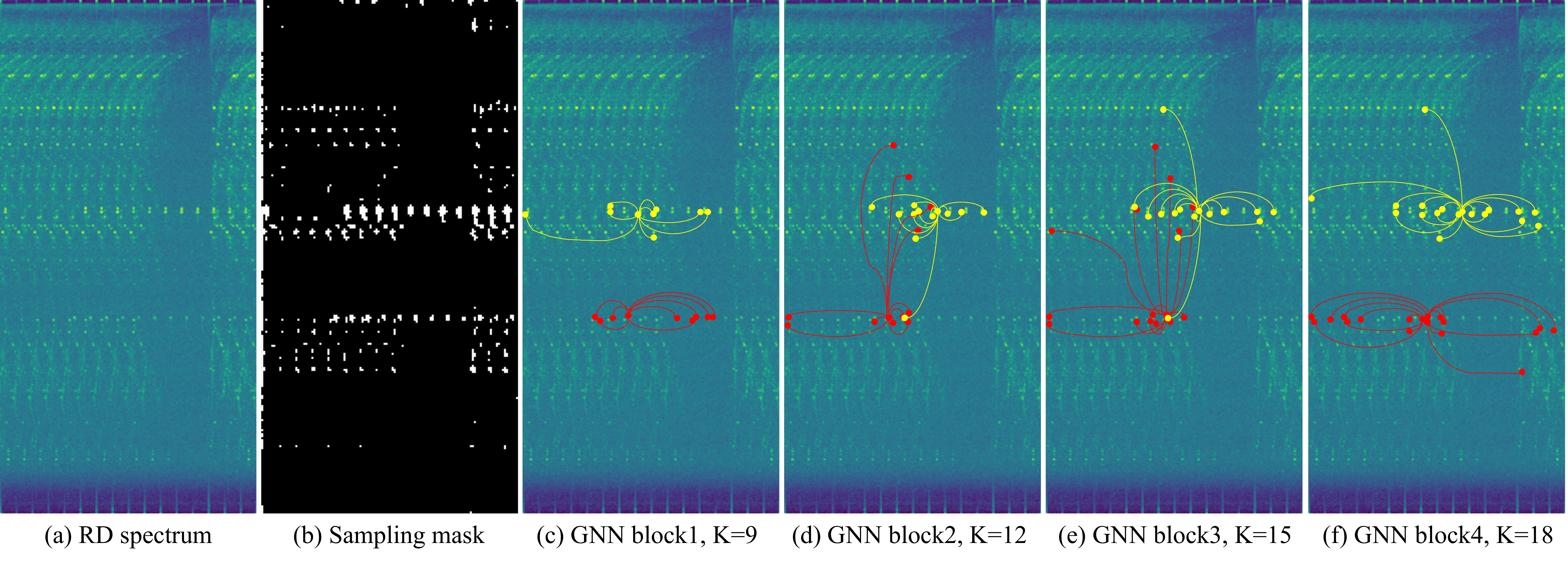}

    \caption{Dynamic edges built by GNN blocks with positional encoding: (a) RD spectrum, (b) dynamic sampling mask, and (c)-(f) dynamic edges.} 
    \label{fig:append_neighbors}
\end{figure}

In our two-branch backbone, the GNN branch captures global neighbor information, and the SCNN branch is responsible for local neighbor information. Therefore, we choose not to include any positional encoding in the GNN blocks to allow the GNN branch to focus solely on global exploration. Table \ref{append:pos_encoding} compares the results of our SparseRadNet with and without positional encoding. Fig. \hyperref[fig:append_neighbors]{5c-f} present examples of dynamic edges built by GNN blocks with positional encoding added. Adding positional encoding makes the dynamic edges dedicate more to the repetition of object signatures. However, this adjustment does not lead to any noticeable difference in the overall accuracy. This is due to the SCNN branch already efficiently handling local information.

\begin{table}[tbp]\centering
    \caption{Ablation study on positional encoding in GNN blocks. }
    \label{append:pos_encoding}
    \resizebox{0.65\textwidth}{!}{
\begin{tabular}{c|cccccc|cc}
\hline
SparseRadNet              & F1$\uparrow$ & AP$\uparrow$ & AR$\uparrow$ & RE$\downarrow$ & AE$\downarrow$ & mIoU$\uparrow$  & FLOPs$\downarrow$ & Params$\downarrow$ \\ \hline
with pos. encoding & 93.79 & 96.80 & 90.97 & 0.13 & 0.10 & 79.22 & 275G  &  6.9M      \\
w/o pos. encoding & 93.84 & 96.00 & 91.78 & 0.13 & 0.10 & 78.48 & 259G &  6.9M \\ \hline
\end{tabular}
    }
\end{table}


\subsubsection{GNN branch.}

in \cref{append:rebuttal_GNN}, We add additional ablation studies on the GNN branch in terms of the number of blocks, neighbors and nodes. When choosing the number of GNN blocks, the model capability should be prioritized. The number of GNN neighbors should be sufficient to cover repeated object signatures while avoiding the generation of dense graphs. Providing enough samples is crucial for perception. However, an overly large sample size incurs radar noise and burden to GNN. Adding positional encoding mitigates the impact of noise when the node count is large, while for 4000 samples no noticeable difference (\cref{append:pos_encoding}). Compared to static graphs, dynamic GNNs integrate global information and collaborate with the SCNN branch that focuses on local features, demonstrating higher accuracy.

\begin{table}[htbp]\centering
    \caption{Ablation studies on the GNN branch. ``*'': result with positional encoding in GNN blocks.}
    \label{append:rebuttal_GNN}
    \resizebox{0.80\textwidth}{!}{
\begin{tabular}{c|ccc|ccc|cccccc|c}
\hline
\multirow{2}{*}{Metric} & \multicolumn{3}{c|}{\#GNN blocks} & \multicolumn{3}{c|}{\#neighbors} & \multicolumn{6}{c|}{\#nodes (samples) in GNN} & static \\ \cline{2-13}
                        & 2         & 4         & 6         & 6\shorttextrightarrow12      & 9\shorttextrightarrow18      & 12\shorttextrightarrow24    & 1000  & 3000  & 4000  & 6000  & 8000  & 8000* & graph  \\ \hline
F1$\uparrow$            & 93.65     & \underline{93.84}     & 93.36     & 93.02     & \underline{93.84}     & 92.96    & 92.00 & 92.90 & \underline{93.84} & 92.61 & 92.23 & 93.14 & 92.80  \\ \hline
mIoU$\uparrow$          & 79.12     & \underline{78.48}     & 78.49     & 78.61     & \underline{78.48}     & 78.91    & 79.46 & 78.05 & \underline{78.48} & 78.96 & 79.23 & 79.34 & 78.93  \\ \hline
\end{tabular}
    }
\end{table}

\begin{table}[htbp]\centering
    \caption{Ablation study on different fusion modules. AAF denotes axial attentive fusion layers.}
    \label{append:AAF}
    \resizebox{0.5\textwidth}{!}{
\begin{tabular}{c|cccccc}
\hline
SparseRadNet              & F1$\uparrow$ & AP$\uparrow$ & AR$\uparrow$ & RE$\downarrow$ & AE$\downarrow$ & mIoU$\uparrow$\\ \hline
Attentive Addition & 91.71 & 95.43 & 88.26 & 0.13 & 0.10 & 78.02 \\
Addition + CNNs & 92.05 & 96.12 & 88.31 & 0.12 & 0.10 & 77.62 \\
Addition + 4AAF & 93.52 & 96.53 & 90.69 & 0.13 & 0.10 & \textbf{78.57} \\
Addition + 8AAF & \textbf{93.84} & 96.00 & 91.78 & 0.13 & 0.10 & 78.48 \\ \hline
\end{tabular}
    }
\end{table}

\subsubsection{Attentive Fusion.}
Table \ref{append:AAF} compares the model performances with different fusion modules. With only spatial attentive addition, the global and local feature maps from the two-branch backbone are not fully exploited. By applying CNN layers in the fusion, the accuracy increases by 0.3 points. However,  the CNN layers are limited by their local receptive field, preventing them from effectively integrating all relevant global and local features on the feature map. Axial attentive fusion layers incorporate prior knowledge within their context window and address the misalignment of object signatures resulting from different processing in the two branches, thus leading to a boost in accuracy.

\section{Runtime Analysis} \label{append_runtime}

We test the runtime of our SparseRadNet on a GeForce RTX 3090 GPU. The implementation of our SparseRadNet is based on the spconv library \cite{spconv}, which offers runtime acceleration by using half precision (FP16). With this acceleration enabled for those SCNN layers in our model, the inference time is around 61 milliseconds in average, including data loading, network forward pass, and post-processing. The RADIal dataset is recorded by an FMCW radar operating at a frame rate of 5 FPS \cite{FFT-RadNet}. As a result, our model can achieve real-time performance on a GeForce RTX 3090 GPU.

However, we observe that during each network forward pass, approximately 20 to 30 milliseconds are spent on converting dense tensors to sparse tensors. This could be remedied by a more efficient conversion process.

\end{document}